%% file: arxiv.tex
\documentclass[10pt,letterpaper]{article}
\usepackage{kernelzero-arxiv}
\usepackage{wrapfig}
\usepackage{fontawesome5}
\usepackage{etoc}
\usepackage{needspace}
\newcommand{\kzWrapOverview}{1}
\newcommand{\kzWrapAblation}{1}
\usepackage{algorithm,algorithmic}
\usepackage{listings}
\usepackage{booktabs,multirow,makecell}
\usepackage{tikz}
\usetikzlibrary{arrows.meta}
\usepackage{pifont,subcaption,ragged2e,xstring}
\input{appendix/box_styles}
\BeforeBeginEnvironment{prompt}{\tcbset{breakable}}
\AfterEndEnvironment{prompt}{\tcbset{unbreakable}}
\input{appendix/arxiv_box_styles}
\title{\textcolor{kzBlue}{KernelZero}: Co-Evolving Proposer and Coder for Continuously Improved GPU Kernel Generation}
\author{
    \mbox{Changxin Ke\textsuperscript{\rm 1,\rm 2}},
    \mbox{Rui Zhang\textsuperscript{\rm 1,\rm 2}},
    \mbox{Zixiang Fang\textsuperscript{\rm 1,\rm 2}},
    \mbox{Zhenghong Li\textsuperscript{\rm 1,\rm 2}},
    \mbox{Yuanbo Wen\textsuperscript{\rm 1,\rm 2}},\\
    \mbox{Jiaming Guo\textsuperscript{\rm 1,\rm 2}},
    \mbox{Ling Li\textsuperscript{\rm 2,\rm 3}},
    \mbox{Qi Guo\textsuperscript{\rm 1,\rm 2}},
    \mbox{Yunji Chen\textsuperscript{\rm 4,\rm 1,\rm 2}\corresponding}
}
\affiliations{
    \textsuperscript{\rm 1}State Key Lab of Processors, Institute of Computing Technology, CAS\\
    \textsuperscript{\rm 2}University of Chinese Academy of Sciences\\
    \textsuperscript{\rm 3}Intelligent Software Research Center, Institute of Software, CAS\\
    \textsuperscript{\rm 4}Institute of AI for Industries, Chinese Academy of Sciences
}

\begin{document}
\etocdepthtag.toc{main}
\maketitle
\begin{kztitlepage}
\kzauthors
{\centering\setlength{\parindent}{0pt}\small
\href{https://huggingface.co/collections/kcxain/kernelzero}{%
\raisebox{-0.22em}{\kzOriginalIncludegraphics[height=1.15em]{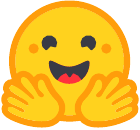}}\enspace \textbf{Model}}
\qquad
\href{https://github.com/kcxain/KernelZero}{\raisebox{-0.22em}{\kzOriginalIncludegraphics[height=1.15em]{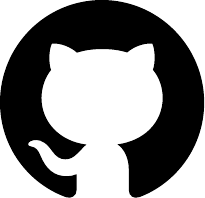}}\enspace \textbf{Code}}
\par}
\bigskip
\begin{abstract}
High-performance GPU kernels are essential to modern machine learning systems, yet automatically generating kernels that are both correct and efficient remains challenging. Existing LLM-based approaches face two major limitations: the scarcity of high-quality training data aligned with the model's current capabilities, and the inherent trade-off between kernel correctness and performance. To address these challenges, we propose \textbf{KernelZero}, a co-evolution framework that continuously improves GPU kernel generation through two specialized models: a Proposer that generates Torch modules from API sets and a Coder that translates them into CUDA or Triton kernels. KernelZero uses a frontier-driven module generation mechanism to continuously produce capability-aligned training modules based on the Coder's current weaknesses. It further introduces Correctness-Aware Group Relative Policy Optimization (CA-GRPO), which optimizes performance only after correctness becomes sufficiently reliable. By alternating the optimization of the Proposer and Coder, KernelZero forms an automatic curriculum that enables targeted and training-efficient capability improvement. Empirically, KernelZero-7B surpasses Claude-4.5-Sonnet on CUDA and DeepSeek-V4-Pro on Triton. On KernelBench Level 1 and 2, it achieves CUDA pass@1 scores of $75.8\%$ and $69.6\%$, respectively, with pass@10 reaching $100\%$ and $97\%$. On Triton, it achieves pass@1 scores of $77.2\%$ and $72.5\%$, respectively.
\end{abstract}
\smallskip
{\setlength{\parindent}{0pt}\fontsize{8.5}{10}\selectfont
\rule{0.30\linewidth}{0.3pt}\par
\textsuperscript{*}Corresponding author.
\par}
\end{kztitlepage}
\suppressfloats[t]

\input{section/intro}
\input{section/method}
\input{section/experiment}
\input{section/related}

\input{section/conclusion}

\bibliographystyle{plainnat}
\bibliography{references}

\appendix
\clearpage
\etocdepthtag.toc{appendix}
\etocsettagdepth{main}{none}
\etocsettagdepth{appendix}{subsubsection}
\etocsettocdepth{subsubsection}
\renewcommand{\contentsname}{Appendix Contents}
\tableofcontents
\clearpage
\input{appendix/evaluate_details}
\input{appendix/grpo}
\input{appendix/hyperparams}
\raggedbottom
\input{appendix/prompts}

\input{appendix/case_proposer}
\input{appendix/case_backend}

\end{document}

%% file: appendix/box_styles.tex
\definecolor{kzPromptBack}{RGB}{246,248,251}
\definecolor{kzPromptFrame}{RGB}{126,143,160}
\definecolor{kzPromptTitleBack}{RGB}{224,231,238}
\definecolor{kzPromptTitleText}{RGB}{48,65,82}
\definecolor{kzCodeBack}{RGB}{249,249,248}
\definecolor{kzCodeFrame}{RGB}{158,164,169}
\definecolor{kzCodeKeyword}{RGB}{55,78,101}
\definecolor{kzCodeComment}{RGB}{91,111,96}
\definecolor{kzCodeLabelColor}{RGB}{55,72,89}

\lstdefinestyle{kzcode}{
  language=Python,
  basicstyle=\ttfamily\fontsize{7.5}{8.5}\selectfont,
  keywordstyle=\color{kzCodeKeyword}\bfseries,
  commentstyle=\color{kzCodeComment}\itshape,
  stringstyle=\color{black},
  backgroundcolor=\color{kzCodeBack},
  rulecolor=\color{kzCodeFrame},
  frame=lrb,
  framerule=0.4pt,
  framesep=4pt,
  numbers=none,
  tabsize=2,
  breaklines=true,
  breakatwhitespace=false,
  breakautoindent=true,
  breakindent=1em,
  columns=fullflexible,
  keepspaces=true,
  showstringspaces=false,
  upquote=true,
  aboveskip=0pt,
  belowskip=5pt,
  xleftmargin=4pt,
  xrightmargin=4pt,
  framexleftmargin=0pt,
  framexrightmargin=0pt,
  linewidth=\linewidth
}

\newcommand{\kzCodeLabel}[1]{%
  \StrGobbleRight{#1}{1}[\kzCodeTitleText]%
  \par\smallskip
  \begin{tcolorbox}[
    width=\linewidth,
    enhanced jigsaw,
    colback=kzPromptTitleBack,
    colframe=kzCodeFrame,
    boxrule=0.4pt,
    arc=1pt,
    outer arc=1pt,
    sharp corners=south,
    left=1.6mm,
    right=1.6mm,
    top=0.25mm,
    bottom=0.25mm,
    before skip=0pt,
    after skip=0pt
  ]
  {\bfseries\fontsize{8}{9}\selectfont\color{kzCodeLabelColor}\kzCodeTitleText}
  \end{tcolorbox}\nobreak
}

\newtcolorbox{prompt}[1][]{
  width=\linewidth,
  enhanced jigsaw,
  colback=kzPromptBack,
  colframe=kzPromptFrame,
  colbacktitle=kzPromptTitleBack,
  coltitle=kzPromptTitleText,
  boxrule=0.45pt,
  arc=1pt,
  outer arc=1pt,
  left=1.6mm,
  right=1.6mm,
  top=1.3mm,
  bottom=1.3mm,
  before skip=5pt,
  after skip=7pt,
  title={#1},
  fonttitle=\bfseries\footnotesize,
  fontupper=\footnotesize\linespread{1.12}\selectfont,
  before upper={
    \RaggedRight
    \sloppy
    \emergencystretch=2em
    \setlength{\parindent}{0pt}
    \setlength{\parskip}{0.45\baselineskip}
  }
}

%% file: appendix/arxiv_box_styles.tex
\renewtcolorbox{prompt}[1][]{
  report panel,
  title={#1},
  fontupper=\rmfamily\fontsize{9}{11}\selectfont,
  before upper={\RaggedRight\sloppy\emergencystretch=2em
    \setlength{\parindent}{0pt}\setlength{\parskip}{5pt}}
}
\def\kzCodeTitleText{}
\renewcommand{\kzCodeLabel}[1]{\StrGobbleRight{#1}{1}[\kzCodeTitleText]}
\lstdefinestyle{kzcode}{
  language=Python,basicstyle=\ttfamily\fontsize{7.5}{9}\selectfont,
  keywordstyle=\color{kzBlue}\bfseries,
  commentstyle=\color{kzCodeComment},stringstyle=\color{black},
  backgroundcolor={},frame=none,numbers=none,tabsize=2,
  breaklines=true,breakatwhitespace=false,breakautoindent=true,breakindent=1em,
  columns=fullflexible,keepspaces=true,showstringspaces=false,upquote=true,
  aboveskip=0pt,belowskip=0pt,xleftmargin=0pt,xrightmargin=0pt,
  linewidth=\linewidth
}
\tcolorboxenvironment{lstlisting}{
  report panel,
  title=\kzCodeTitleText,fonttitle=\rmfamily\bfseries\small,
  code={\ifdefempty{\kzCodeTitleText}{\tcbset{notitle}}{}}
}
\AfterEndEnvironment{lstlisting}{\gdef\kzCodeTitleText{}}

%% file: section/intro.tex
\section{Introduction}

\ifdefined\kzWrapOverview\else
\begin{figure}[t]
  \begin{center}
\centerline{\includegraphics[width=1.0\linewidth]{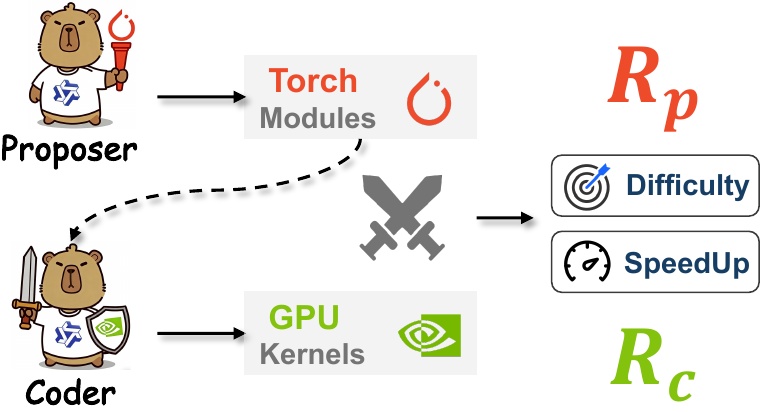}}
    \caption{
Overview of \textbf{KernelZero}, a co-evolution framework for kernel generation. The system consists of two models with explicit interfaces: a \emph{Proposer} that takes a set of Torch APIs as input and generates a runnable Torch module, and a \emph{Coder} that translates the module into an optimized CUDA or Triton kernel.}
    \label{overview_1}
  \end{center}
  \vspace{-20pt}
\end{figure}
\fi

High-performance GPU kernels are critical to modern machine learning systems. However, developing them remains costly and time-consuming, demanding knowledge of domain-specific languages such as CUDA~\cite{cuda}, BangC~\cite{bangc}, and Triton~\cite{triton}, as well as expertise in computer architectures. With the development of LLMs, using LLMs for GPU kernel generation has become possible. 
For example, recent studies including Kevin~\citep{kevin} and Dr.Kernel~\citep{drkernel} have demonstrated the capability of LLMs to generate high-performance GPU kernels on systematic evaluation frameworks such as KernelBench~\citep{kernelbench} and FlashInfer-Bench~\citep{flashinferbench}.
However, existing results show that even with extensive SFT and RL training, LLMs still face challenges in generating correct and high-performance GPU kernels. 
For example, despite being trained on 128 H100 GPUs, cudaLLM~\citep{cudaLLM} achieves only 81\% pass@10 on KernelBench Level 1.
This leads to our investigation:
\ifdefined\kzResearchQuestion
\begin{kzResearchQuestion}
How can we train models in a targeted way to efficiently improve their GPU kernel generation performance?
\end{kzResearchQuestion}
\else
\textit{How can we train models in a targeted way to efficiently improve their GPU kernel generation performance?}
\fi

Despite existing progress, training LLMs for GPU kernel generation faces two fundamental challenges. First, high-quality data that is aligned with and sensitive to the model’s current capabilities is scarce. As a result, training either fails to converge or incurs extremely high costs. 
Second, there exists a clear trade-off between correctness and performance. Optimizing solely for correctness often leads to conservative kernels with limited speedups, while aggressive performance optimizations significantly increase the risk of incorrect results. 
These challenges make it difficult to design a training paradigm that is both cost-effective and capable of improving GPU kernel generation correctness and performance.

\ifdefined\kzWrapOverview
\begin{wrapfigure}{R}{0.43\textwidth}
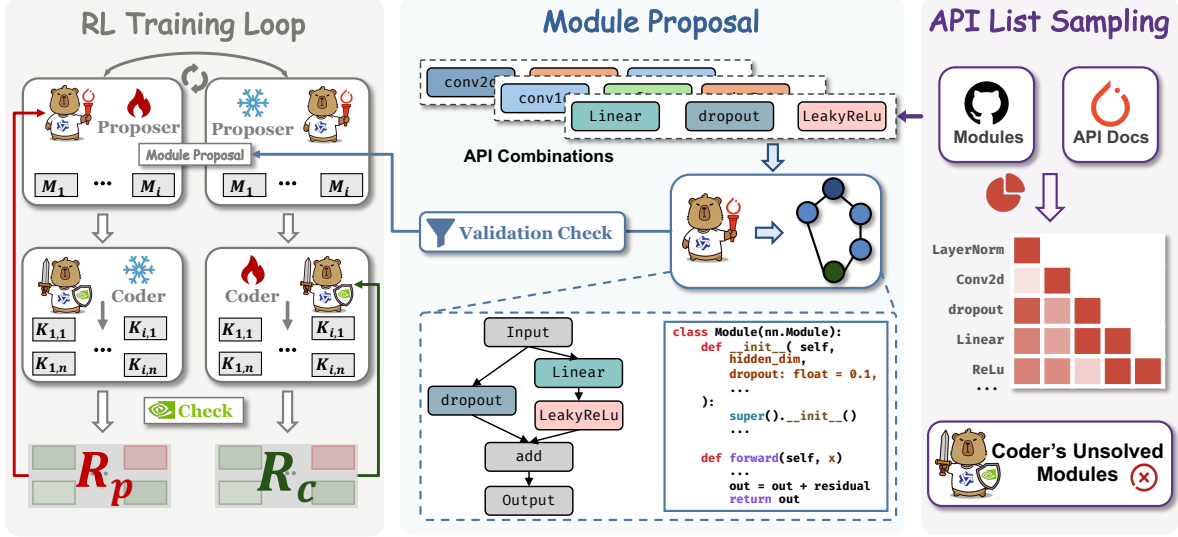

\centering
\captionsetup{width=\linewidth}
\kzOriginalIncludegraphics[width=\linewidth]{figures/overview_1.pdf}
\caption{Overview of \textbf{KernelZero}, a co-evolution framework for kernel generation. The system consists of two models with explicit interfaces: a \emph{Proposer} that takes a set of Torch APIs as input and generates a runnable Torch module, and a \emph{Coder} that translates the module into an optimized CUDA or Triton kernel.}
\label{overview_1}
\end{wrapfigure}
\fi

In this paper, we propose \textbf{KernelZero}, a co-evolving framework that enables the continuous advancement of GPU kernel generation with two specialized models.
Our insight is that a continuous training paradigm enables the model to dynamically identify and learn from the most capability-aligned samples at each iteration, thereby allowing more targeted and efficient improvement.
KernelZero consists of a Proposer that generates Torch modules from API sets and a Coder that translates these modules into CUDA or Triton kernels, as shown in Figure~\ref{overview_1}. 
To address the lack of capability-aligned data, we introduce a frontier-driven module generation mechanism that adaptively aligns the modules generated by the Proposer with the Coder's current capabilities. 
Specifically, the mechanism extracts API sets from modules that the Coder fails to solve at each iteration to construct targeted inputs for the Proposer, while correctness-based feedback guides the Proposer to generate modules near the Coder's capability frontier, avoiding modules that are either trivial or overly difficult. 
This module generation mechanism produces a continuous stream of high-quality, adaptive training data without manual annotation.
To resolve the trade-off between correctness and performance, we propose Correctness-Aware Group Relative Policy Optimization (CA-GRPO) for Coder training. 
This method separates correctness and performance signals through a group-level gating mechanism, ensuring that performance optimization is activated only when correctness becomes sufficiently reliable. 
The Proposer and the Coder are updated in turn using their respective rewards. 
This co-evolution process forms an automatic curriculum that continuously expands the model's ability to generate kernels with higher correctness and performance while requiring fewer training steps.

We evaluate KernelZero on both CUDA and Triton generation. Despite having only 7B parameters, it surpasses Claude-4.5-Sonnet on CUDA and DeepSeek-V4-Pro on Triton. KernelZero achieves $75.8\%$/$69.6\%$ pass@1 on CUDA Levels~1/2, with pass@10 reaching $100\%$/$97\%$, and $77.2\%$/$72.5\%$ pass@1 on Triton Levels~1/2. Continuously updating the Proposer yields pass@1 gains of $0.9$/$5.2$ percentage points on CUDA and $2.3$/$4.2$ points on Triton. It also improves fast$_1$@1 by $1.7$/$0.4$ points on CUDA and $1.2$/$3.5$ points on Triton.

%% file: section/method.tex
\section{Method}

\subsection{KernelZero Framework}

KernelZero is an end-to-end framework that formulates kernel generation as the co-evolution of two models with explicitly defined interfaces: a \textit{Proposer} $P$ and a \textit{Coder} $C$.
The Proposer takes a set of Torch APIs $a \in \mathcal{A}$ as input, and generates a compositional Torch module
$
m \sim \pi_P(\cdot \mid a; \theta_P), m \in \mathcal{M},
$
where $\mathcal{M}$ denotes the space of valid and runnable Torch modules. 
The generated module $m$ defines a structured computation graph that serves as a high-level specification.
Given a module $m$, the Coder subsequently produces a corresponding CUDA or Triton kernel
$
k \sim \pi_C(\cdot \mid m; \theta_C),  k \in \mathcal{K},
$
where $\mathcal{K}$ is the space of executable kernels. The kernel $k$ is required to be functionally correct with respect to $m$ and optimized for hardware efficiency.
This establishes a two-stage mapping
$
a \xrightarrow{P} m \xrightarrow{C} k,
$
where $P$ explores the space of operator compositions and $C$ realizes them as efficient low-level implementations.

\begin{figure*}[t]
  \begin{center}
\centerline{\includegraphics[width=\linewidth]{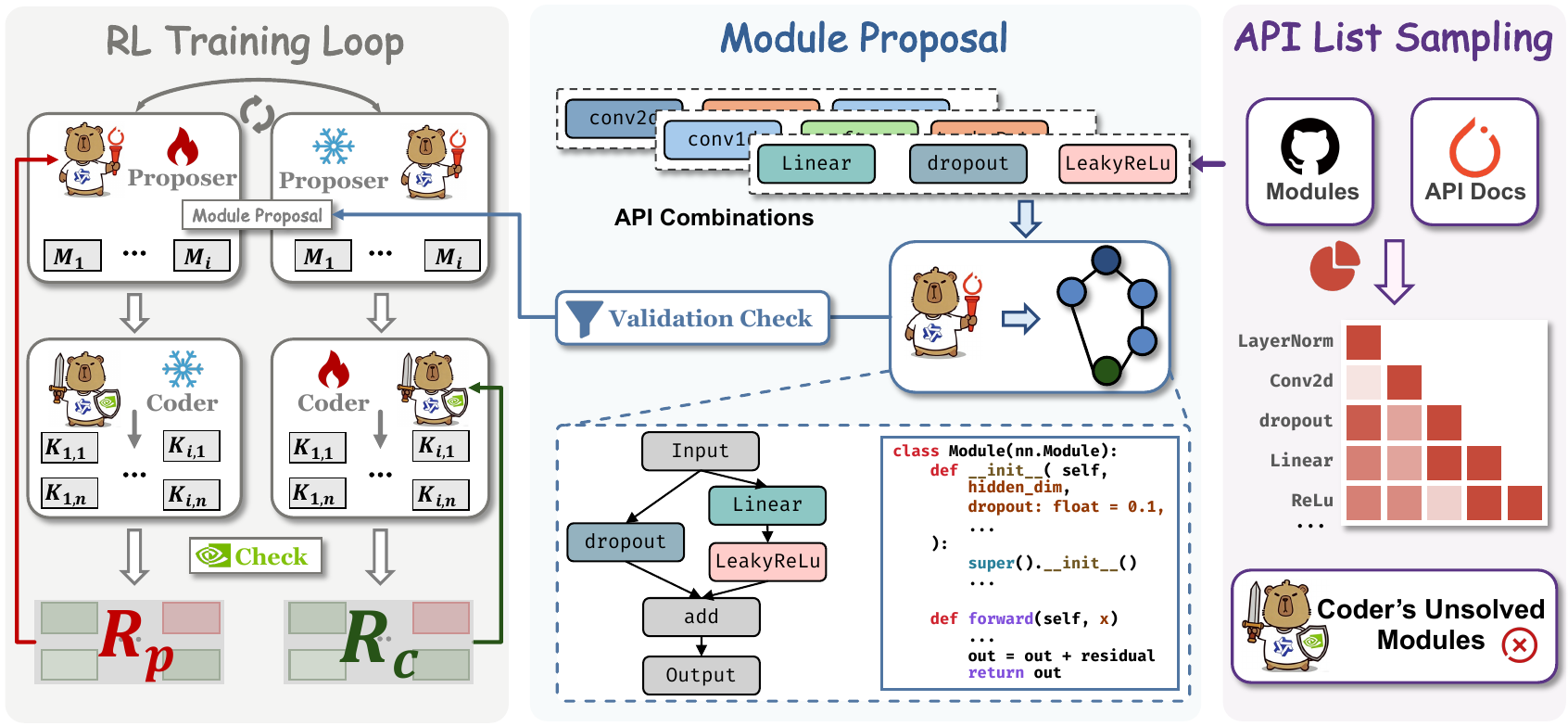}}
    \caption{
Overview of \textbf{KernelZero}, a co-evolution framework for kernel generation. The system consists of two models with explicit interfaces: a \emph{Proposer} that takes a set of Torch APIs as input and generates a runnable Torch module, and a \emph{Coder} that translates the module into an optimized CUDA or Triton kernel.}
    \label{overview}
  \end{center}
  \vspace{-10pt}
\end{figure*}

The whole KernelZero framework adopts an alternating optimization strategy for co-evolving the two models. The Proposer and the Coder are updated in turn: each model is optimized using its own reward ($r_P$ or $r_C$) via GRPO while keeping the other fixed. This alternating procedure enables stable co-evolution, where the Proposer progressively expands the module distribution towards the frontier of the Coder's capability, and the Coder improves its kernel generation quality by adapting to the induced distribution.
An overview of KernelZero is illustrated in Figure~\ref{overview}. We describe the detailed training procedures for each role in the following subsections.

\subsection{Proposer: Proposing Frontier Torch Modules}


The goal of the Proposer is to explore the space of operator compositions and generate high-quality modules at the frontier of the Coder's capability, thereby progressively expanding the Coder's skill boundary through targeted challenges.
To achieve this goal, the modules generated from the Proposer should be expressive, reliable and frontier-oriented. 
Thus, the Proposer applies API-list sampling before module generation to encourage realistic and diverse operator combinations, followed by validation checks for correctness and usability. We also introduce a carefully designed Frontier Reward for the Proposer.

\paragraph{API List Sampling.}
The generated modules should be expressive, meaning that they should exhibit structural patterns similar to realistic Torch API compositions while maintaining sufficient diversity to cover a broad range of operator combinations. This diversity improves the Coder's generalization ability.
To this end, we construct an API co-occurrence distribution from real-world Torch modules and sample a subset $a \subset \mathcal{A}$ from this distribution as input to the Proposer. 
This process favors realistic API compositions while retaining sufficient diversity for exploration.
Given a sampled API set $a$, the Proposer generates a Torch module $m$ that defines a structured computation graph composed of the provided APIs.


\paragraph{Validation Check.}
The generated modules should also be reliable, meaning that each module satisfies strict correctness and usability constraints: it must be functionally valid, executable without runtime errors, and numerically stable.
Thus, each proposed module is subjected to a multi-stage validation procedure to ensure its correctness and usability, including the following three stages.
\textbf{(1) AST Check.} We first perform static analysis using an abstract syntax tree (AST) checker to verify semantic consistency, i.e., that the computation graph defined in the generated module matches the sampled API set $a$.
\textbf{(2) Runnable Check.} We then validate whether the module is executable. Specifically, each generated module includes two auxiliary methods, \texttt{get\_init\_inputs} and \texttt{get\_inputs}, following the format of KernelBench~\cite{kernelbench}, where the former initializes input tensors with appropriate shapes and types, while the latter produces randomized inputs for testing.
We invoke the \texttt{get\_init\_inputs} and \texttt{get\_inputs} methods to construct valid inputs and execute the module on a CUDA device. Modules that fail to compile or raise runtime errors are discarded. 
\textbf{(3) Numerical Stability.} Finally, we evaluate the numerical stability of the module by running it on randomized inputs. Any sample that produces invalid values such as NaN or Inf is filtered out.
Only modules that pass all three validation steps are considered valid. This validation pipeline ensures that the Proposer generates semantically consistent, executable, and numerically stable modules, enabling reliable training of the Coder.

\paragraph{Frontier Reward.}

Finally, to be maximally informative for improving the Coder, generated modules should be frontier-oriented, i.e., at the frontier of the Coder's capability. The Proposer is therefore encouraged to generate modules near the Coder's current capability boundary.
To this end, we introduce \textit{Frontier-Oriented Group Relative Policy Optimization (FO-GRPO)} with a carefully designed Frontier Reward for the Proposer.
Given a proposed module $m$, we sample a rollout group $\mathcal{G} = \{k_1, \dots, k_K\}$ from the Coder. For each kernel $k_i$, we compute the average correctness which reflects the current Coder's capability on module $m$:
\[
\bar{l} = \frac{1}{K} \sum_{i=1}^{K} \mathbf{1}[\mathrm{Corr}(m, k_i)=1].
\]
The Frontier Reward for Proposer depends only on the average correctness of the Coder and is maximized when the success rate is close to $0.5$:
\[
R_p(m)=
\begin{cases}
1 - 2\big| \bar{l} - 0.5 \big|, & m \text{ is valid}, \\
\rho, & m \text{ is invalid},
\end{cases}
\]
where $\rho < 0$ penalizes invalid modules.
This formulation encourages the Proposer to generate modules that are neither trivial nor impossible~\cite{r-zero}. The reward is highest when approximately half of the sampled kernels are correct, which corresponds to the frontier of the Coder's capability.

\subsection{Coder: Generating High-Performance Kernels}\label{sec:coder}

The Coder aims to translate a Torch module into a CUDA or Triton kernel that serves as a low-level implementation of the input Torch module.
The generated kernel is required to satisfy functional correctness while achieving high speedup over the corresponding Torch module.
To this end, we propose a correctness-aware reward to balance correctness and performance.
Kernel implementation also requires substantial domain-specific knowledge. Before reinforcement learning, we therefore introduce compiler-inspired cold-start distillation to equip the Coder with optimization principles from the earliest stage.

\paragraph{Cold Start Distillation}
Modern GPU compilers and expert programmers typically rely on a set of fundamental optimization principles to transform a naive algorithm into a complex, high-performance GPU kernel implementation.
Embedding these compiler-inspired principles during cold-start distillation helps the Coder internalize optimization patterns and benefits subsequent reinforcement learning.
In this way, we summarize some optimization principles that are essential for bridging the gap between high-level operator specifications and low-level hardware efficiency. We utilize these principles to prompt a teacher model to reason about and apply optimization strategies during CUDA and Triton kernel distillation.

Specifically, we summarize four optimization principles that are both fundamental and complementary in GPU kernel design~\citep{qimeng-kernel,qimeng_gemm}:
\begin{itemize}
    \item \textbf{Tiling} partitions input tensors into block-level and warp-level tiles that fit on-chip memory, improving data reuse and reducing expensive global-memory access; it also enables efficient block/warp reductions for collective operators~\citep{tiling}.
    \item \textbf{Fusion} merges producer--consumer operator chains into a single kernel so intermediate tensors stay in registers or shared memory, reducing kernel-launch overhead, DRAM traffic, and synchronization cost~\citep{fusion}.
    \item \textbf{Pipeline} organizes computation into staged execution that overlaps memory movement (e.g., global-to-shared prefetch) with arithmetic, thereby hiding memory latency and increasing sustained SM utilization~\citep{pipeline}.
    \item \textbf{Reordering} rearranges loop/order-of-operations and thread/data mapping to improve coalesced access and temporal reuse, which strengthens cache/shared-memory locality and alleviates bank conflicts and divergence~\citep{reordering}.
\end{itemize}

These principles provide a structured taxonomy for reasoning about kernel efficiency, covering data reuse, memory latency hiding, execution overlap, and access pattern optimization. Each principle addresses a distinct bottleneck in the GPU memory hierarchy and execution model.
By embedding these four optimization principles into the cold start training set, the Coder adheres to established GPU optimization heuristics.

\paragraph{Correctness-Aware Group Relative Policy Optimization}
Kernel generation involves a fundamental trade-off between correctness and performance. Optimizing only for correctness often leads to conservative implementations with limited speedup. In contrast, focusing excessively on performance can break functional equivalence. This trade-off makes direct reward design unstable, especially in early training.
To address this issue, we propose Correctness-Aware Group Relative Policy Optimization (CA-GRPO) to separate correctness and performance signals through a group-level mechanism, so that performance optimization is activated only when correctness becomes sufficiently reliable.

Given a module $m$, the Coder samples a rollout group $\mathcal{G} = \{k_1, \dots, k_n\}$ of candidate kernels. We compute the group-level correctness:
\[
\mathrm{Acc}_{\mathcal{G}} = \frac{1}{|\mathcal{G}|} \sum_{i=1}^n \mathbf{1}[\mathrm{Corr}(m, k_i)=1],
\]
and define a group correctness threshold function $\mathcal{T}(\mathcal{G})$ where the threshold $\alpha$ determines when the optimization process transitions from correctness to performance:
\[
\mathcal{T}(\mathcal{G}) =
\begin{cases}
1, & \text{if } \mathrm{Acc}_{\mathcal{G}} \ge \alpha, \\
0, & \text{otherwise}.
\end{cases}
\]

When group correctness reaches the threshold $\alpha$, correct kernels receive the performance reward.
Let $\mathcal{G}_c \subset \mathcal{G}$ denote the subset of correct kernels. For each $k_i \in \mathcal{G}_c$, we define a normalized speedup score as:
\[
\mathcal{S}_i^{\mathcal{G}} =
\begin{cases}
\frac{
\mathbf{S}(k_i) - \min(\mathbf{S}(\mathcal{G}_c))
}{
\max(\mathbf{S}(\mathcal{G}_c)) - \min(\mathbf{S}(\mathcal{G}_c))
}, & \max(\mathbf{S}(\mathcal{G}_c)) > \min(\mathbf{S}(\mathcal{G}_c)), \\
0, & \text{otherwise},
\end{cases}
\]
where $\mathbf{S}(k_i)$ denotes the measured speedup. This normalization preserves the relative ordering of performance within the group and bounds the scale to a fixed range. When all correct kernels have the same measured speedup, their normalized speedup scores are set to zero.
Finally, the proposed CA-GRPO is performed with a correctness-aware reward to ensure that correctness is learned first, while performance optimization is introduced only after the model achieves sufficient reliability.
The correctness-aware reward for each sample $k_i \in \mathcal{G}$ is defined as
\[
\mathcal{R}_i^{\mathcal{G}} =
\begin{cases}
1 + \mathcal{T}(\mathcal{G}) \cdot \beta \cdot \mathcal{S}_i^{\mathcal{G}}, & \text{if } \mathrm{Corr}(m, k_i)=1, \\
0, & \text{otherwise},
\end{cases}
\]
where $\beta$ controls the strength of the performance signal.
With the correctness-aware reward, the training process remains stable, preserving correctness while enabling speedup improvements in later stages.

%% file: section/experiment.tex
\begin{figure*}[!t]
    \centering
    \includegraphics[width=\linewidth]{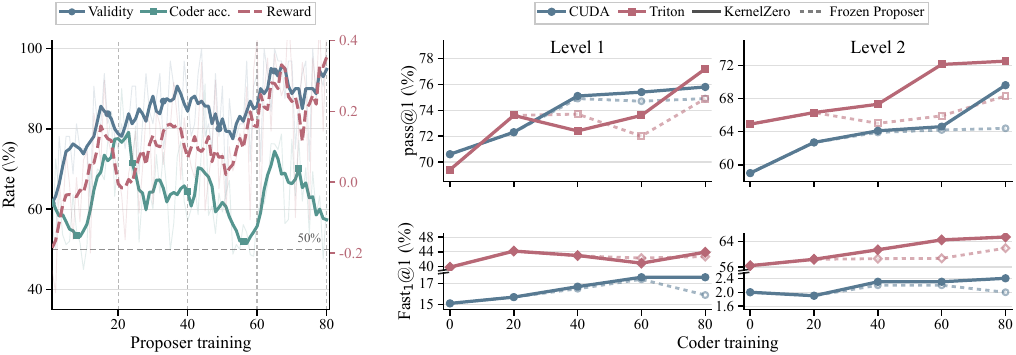}
    \caption{Training dynamics of KernelZero. The left panel shows the validity rate, Coder accuracy, and frontier reward during Proposer training. The right panels report pass@1 and fast$_1$@1 for CUDA and Triton Coder training. The solid curves denote the complete KernelZero training process, while the lighter dotted curves freeze the Proposer from Step~40.}
    \label{fig:proposer_training_dynamics}
\end{figure*}

\ifdefined\kzWrapAblation\Needspace{10\baselineskip}\fi
\section{Experiments}
\subsection{Training and Evaluation Setups}

\subsubsection{Training Details}

Both the Proposer and the Coder are initialized from Qwen2.5-Coder-7B~\citep{qwen25coder}.
For Proposer training, we construct a pool of 4,000 unique API lists from the
co-occurrence distribution, including 2,000 two-API combinations and 2,000
three-API combinations.

\paragraph{Coder Cold Start.}
Before reinforcement learning, the Coder undergoes a cold start distillation phase.
We start from the 71,996 kernel instances publicly released by cudaLLM~\citep{cudaLLM} and re-sample each instance using gpt-oss-120B~\cite{gptoss} as a teacher model.
The teacher is prompted with the four GPU optimization principles described in Section~\ref{sec:coder} (Tiling, Fusion, Pipeline, Reordering) to produce step-by-step chain-of-thought reasoning before writing the final CUDA or Triton kernel.
Each generated kernel is then verified for functional correctness, and only correct samples are retained.
This filtering yields 42,454 CUDA traces and 59,998 Triton traces, on which the corresponding Coder is fine-tuned via supervised learning to internalize both the optimization logic and structured reasoning from the earliest stage.

\paragraph{RL Training Loop.}
After cold start, the Proposer and the Coder are trained alternately via GRPO. In each round, the Proposer is updated for 20 steps using the frontier reward $R_p$, followed by the Coder being updated for 20 steps using the CA-GRPO reward $\mathcal{R}^{\mathcal{G}}$. The two models alternate for a total of 4 rounds. Figure~\ref{fig:proposer_training_dynamics} shows that the validity rate increases during Proposer training, while the Coder accuracy fluctuates near the target frontier and the frontier reward gradually improves.

\ifdefined\kzWrapAblation
\Needspace{0.42\textheight}
\fi
\subsubsection{Evaluation Benchmark and Metrics}
\ifdefined\kzWrapAblation
\begin{wrapfigure}{r}{0.43\textwidth}
\captionsetup{width=\linewidth}
\let\includegraphics\kzOriginalIncludegraphics
\else
\begin{figure}[ht]
\fi
\centering
\includegraphics[width=\linewidth]{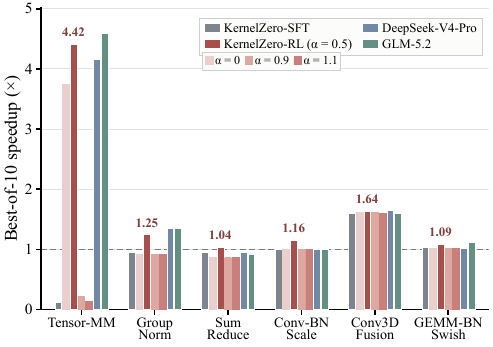}
\caption{\textbf{Ablation of the correctness threshold $\alpha$.}
Each bar reports the speedup of the fastest correct kernel among 10 generations.
KernelZero-SFT denotes the cold start initialization, while KernelZero-RL uses
$\alpha=0.5$. The lighter red bars show the results for $\alpha=0$, $0.9$,
and $1.1$. DeepSeek-V4-Pro and GLM-5.2 are included as external references.
}
\label{fig:reward_ablation_operators}
\ifdefined\kzWrapAblation
\end{wrapfigure}
\else
\end{figure}
\fi

We evaluate KernelZero and baseline models on \textbf{KernelBench}~\citep{kernelbench}, a benchmark covering 250 tasks across three levels of difficulty.
Level 1 contains 100 single-kernel operators (e.g., Convolution), Level 2 contains 100 fusion patterns (e.g., Matmul + Sigmoid).
Each model is given a one-shot prompt to generate a CUDA or Triton kernel based on a PyTorch reference implementation.

We report two categories of metrics. For a task with $n$ sampled kernels, of which $c$ are functionally correct, pass@$k$ is estimated following HumanEval:
\[
\mathrm{pass}@k = 1-\frac{\binom{n-c}{k}}{\binom{n}{k}}.
\]
The metric is averaged over all tasks. fast$_p$@$k$ uses the same estimator, but counts only kernels that are correct and achieve more than $p\times$ speedup as successful samples. For example, fast$_2$@1 reports the fraction of tasks where a single generated kernel is both correct and more than $2\times$ faster than the PyTorch implementation. We additionally report the average number of output tokens, including reasoning and final code, to reflect inference efficiency.

\subsubsection{Baselines}

We compare KernelZero against two categories of baselines, foundation models and specialized models.

\paragraph{Foundation Models.}
We include a broad set of general LLMs applied directly to kernel generation. For CUDA, these baselines cover the Claude and Gemini series, GPT-oss-120B~\citep{gptoss}, GLM-5.1~\citep{glm5}, Qwen3-32B, Qwen3-8B~\citep{qwen3}, Qwen2.5-Coder-32B, and Qwen2.5-Coder-7B~\citep{qwen25coder}. For Triton, we additionally evaluate DeepSeek-V4-Pro~\citep{dpsk-v4} and GLM-5.2.

\paragraph{Specialized Models.}
We compare against models specifically trained for GPU kernel generation. CUDA baselines include cudaLLM-8B~\citep{cudaLLM}, trained on a large-scale supervised dataset using 128 H100 GPUs, and Kevin-32B~\citep{kevin}, which applies multi-turn reinforcement learning to iteratively refine kernel generation. For Triton, we compare with the 8B and 14B variants of Dr.Kernel~\citep{drkernel} under its three-turn setting.

\begin{table*}[!t]
\centering
\definecolor{kzTableCUDA}{HTML}{EAF0F7}
\definecolor{kzTableTriton}{HTML}{EAF2EF}
\definecolor{kzTableOurs}{HTML}{DDE9F5}
\definecolor{kzTableSFT}{HTML}{F1F5F9}
\definecolor{kzTableTritonOurs}{HTML}{DDECE3}
\definecolor{kzTableTritonSFT}{HTML}{F1F6F3}
\small
\begin{tabular}{c | l | c| cccc|cccc}
\toprule
\multirow{2}{*}{\textbf{Kernel}}
& \multirow{2}{*}{\textbf{Model}}
& \multirow{2}{*}{\textbf{Params}}
& \multicolumn{4}{c}{\textbf{Level 1}}
& \multicolumn{4}{c}{\textbf{Level 2}} \\
\cmidrule(lr){4-7}
\cmidrule(lr){8-11}
&
&
&
pass@1
&pass@5
&pass@10
&Tok.
&pass@1
&pass@5
&pass@10
&Tok.
\\
\midrule


\multirow{16}{*}{\textbf{CUDA}}
& \multicolumn{10}{>{\columncolor{kzTableCUDA}}c}{\textit{Foundation Models}}
\\

& Claude-3.5-Sonnet
& --
& 26.7 & 53.42 & 64 & 1.9k
& 5.9 & 21.27 & 32 & 2.6k
\\

& Claude-3.7-Sonnet
& --
& 26.9 & 59.83 & 70 & 3.5k
& 15.6 & 46.03 & 61 & 4.3k
\\

& Claude-4.5-Sonnet
& --
& \textbf{76.4} & \underline{95.35} & 97 & 2.1k
& 66.5 & 90.25 & 93 & 3.1k
\\

& Gemini-2.5-Flash
& --
& 24.2 & 62.76 & 82 & 14.8k
& 13.1 & 44.13 & 62 & 21.2k
\\

& GPT-oss
& 120B
& 35.1 & 83.21 & 94 & 2.9k
& 28.0 & 69.23 & 82 & 4.4k
\\

& GLM-5.1
& 744B
& 22.9 & 44.56 & 53 & 1.4k
& 6.9 & 14.50 & 19 & 1.7k
\\

& Qwen3
& 32B
& 27.8 & 57.91 & 68 & 7.4k
& 12.0 & 35.13 & 46 & 8.9k
\\

& Qwen3
& 8B
& 4.4 & 13.67 & 19 & 4.9k
& 0.7 & 2.92 & 5 & 6.5k
\\

& Qwen2.5-Coder
& 32B
& 16.7 & 33.44 & 43 & \textbf{0.7k}
& 1.2 & 3.90 & 5 & \textbf{1.1k}
\\

& Qwen2.5-Coder
& 7B
& 6.5 & 17.73 & 24 & \underline{0.8k}
& 0.4 & 1.78 & 3 & \underline{1.3k}
\\

& \multicolumn{10}{>{\columncolor{kzTableCUDA}}c}{\textit{Specialized Models}}
\\

& cudaLLM
& 8B
& 72.5 & 79.58 & 81 & 5.3k
& \underline{67.4} & 73.91 & 77 & 8.4k
\\

& Kevin
& 32B
& 42.8 & 72.17 & 82 & 9.2k
& 19.0 & 49.91 & 62 & 13.7k
\\

& \cellcolor{kzTableSFT}\textbf{KernelZero-SFT}
& \cellcolor{kzTableSFT}7B
& \cellcolor{kzTableSFT}70.6 & \cellcolor{kzTableSFT}94.77 & \cellcolor{kzTableSFT}\underline{98} & \cellcolor{kzTableSFT}2.7k
& \cellcolor{kzTableSFT}59 & \cellcolor{kzTableSFT}\underline{90.33} & \cellcolor{kzTableSFT}\underline{94} & \cellcolor{kzTableSFT}4.5k
\\

& \cellcolor{kzTableOurs}\textbf{KernelZero}
& \cellcolor{kzTableOurs}7B
& \cellcolor{kzTableOurs}\underline{75.8} & \cellcolor{kzTableOurs}\textbf{98.87} & \cellcolor{kzTableOurs}\textbf{100} & \cellcolor{kzTableOurs}2.8k
& \cellcolor{kzTableOurs}\textbf{69.6} & \cellcolor{kzTableOurs}\textbf{93.70} & \cellcolor{kzTableOurs}\textbf{97} & \cellcolor{kzTableOurs}4.5k
\\

\midrule


\multirow{11}{*}{\textbf{Triton}}
& \multicolumn{10}{>{\columncolor{kzTableTriton}}c}{\textit{Foundation Models}}
\\

& DeepSeek-V4-Pro
& 1.6T
& 49.7 & 68.06 & 73 & 12.2k
& 63.4 & 88.69 & 91 & 9.9k
\\

& GLM-5.2
& 744B
& 66.7 & 76.33 & 78 & 19.6k
& 60.9 & 67.97 & 70 & 15.8k
\\

& GPT-oss
& 120B
& 55.7 & 88.68 & 95 & 4.7k
& 42.7 & 66.28 & 72 & 4.4k
\\

& Qwen2.5-Coder
& 32B
& 20.3 & 41.98 & 49 & \underline{3.3k}
& 6.8 & 20.52 & 29 & \underline{3.2k}
\\

& Qwen2.5-Coder
& 7B
& 0.3 & 1.28 & 2 & \textbf{1.5k}
& 0.9 & 4.28 & 8 & \textbf{1.7k}
\\

& \multicolumn{10}{>{\columncolor{kzTableTriton}}c}{\textit{Specialized Models}}
\\

& Dr.Kernel (3 turn)
& 8B
& 34.4 & -- & -- & --
& \textbf{73.0} & -- & -- & --
\\

& Dr.Kernel (3 turn)
& 14B
& 30.9 & -- & -- & --
& 57.5 & -- & -- & --
\\

& \cellcolor{kzTableTritonSFT}\textbf{KernelZero-SFT}
& \cellcolor{kzTableTritonSFT}7B
& \cellcolor{kzTableTritonSFT}\underline{69.4} & \cellcolor{kzTableTritonSFT}\underline{96.38} & \cellcolor{kzTableTritonSFT}\underline{98} & \cellcolor{kzTableTritonSFT}4.0k
& \cellcolor{kzTableTritonSFT}64.9 & \cellcolor{kzTableTritonSFT}\underline{89.92} & \cellcolor{kzTableTritonSFT}\underline{94} & \cellcolor{kzTableTritonSFT}3.8k
\\

& \cellcolor{kzTableTritonOurs}\textbf{KernelZero}
& \cellcolor{kzTableTritonOurs}7B
& \cellcolor{kzTableTritonOurs}\textbf{77.2} & \cellcolor{kzTableTritonOurs}\textbf{97.41} & \cellcolor{kzTableTritonOurs}\textbf{99} & \cellcolor{kzTableTritonOurs}3.8k
& \cellcolor{kzTableTritonOurs}\underline{72.5} & \cellcolor{kzTableTritonOurs}\textbf{94.96} & \cellcolor{kzTableTritonOurs}\textbf{98} & \cellcolor{kzTableTritonOurs}3.7k
\\

\bottomrule
\end{tabular}

\caption{
Execution accuracy and token cost of \textbf{CUDA} and \textbf{Triton}
kernel generation on KernelBench.
The best and second-best available results within each backend are
highlighted in \textbf{bold} and \underline{underlined}, respectively.
Lower token cost is better.
}

\label{tab:cuda_triton_pass_tokens}
\end{table*}

\subsection{Main Results}

\begin{figure*}[t]
\centering
\includegraphics[width=\textwidth]{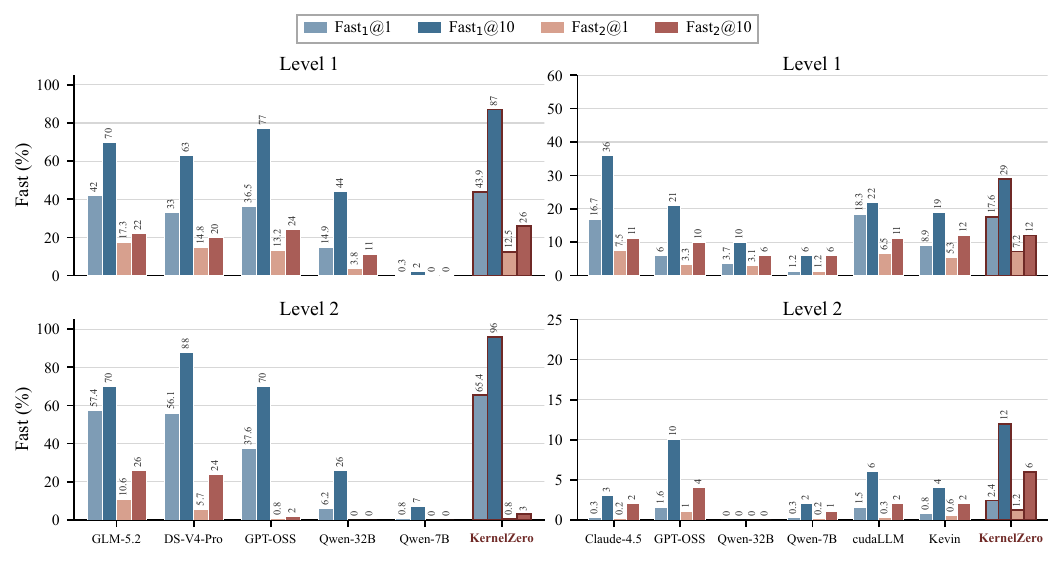}
\caption{
Fast@$k$ results on KernelBench. The left column reports Triton generation
and the right column reports CUDA generation. Fast$_p$@$k$ is the percentage
of tasks for which at least one of $k$ generated kernels is correct and
achieves more than $p\times$ speedup over PyTorch Eager. KernelZero is
compared with representative foundation and specialized models.
}
\label{fig:fast_baseline_comparison}
\end{figure*}

Table~\ref{tab:cuda_triton_pass_tokens} and Figure~\ref{fig:fast_baseline_comparison} report execution accuracy and speedup on KernelBench. KernelZero achieves strong execution accuracy despite its compact model size. For CUDA, it obtains 75.8 pass@1, 98.87 pass@5, and 100 pass@10 on Level 1, and 69.6 pass@1, 93.70 pass@5, and 97 pass@10 on Level 2. For Triton, KernelZero reaches 77.2 pass@1, 97.41 pass@5, and 99 pass@10 on Level 1, and 72.5 pass@1, 94.96 pass@5, and 98 pass@10 on Level 2.

Compared with foundation models, KernelZero is competitive with much larger systems. For CUDA, its Level-1 pass@1 is within 0.6 percentage points of Claude-4.5-Sonnet, while its pass@5 and pass@10 are higher on both levels. For Triton, KernelZero obtains the highest pass@1 on Level 1 and is within 0.5 percentage points of Dr.Kernel-8B on Level 2. It also achieves the highest pass@5 and pass@10 among the Triton models with reported multi-sample results. These results suggest that targeted training for kernel generation can close much of the gap to substantially larger models.

In terms of speedup, KernelZero also produces efficient kernels. On Level 1, it achieves 17.6 fast$_1$@1 and 7.2 fast$_2$@1, outperforming cudaLLM-8B on fast$_2$@1. On Level 2, KernelZero obtains 2.4 fast$_1$@1 and 1.2 fast$_2$@1, exceeding both specialized baselines under the stricter fast$_2$ criterion.

Figure~\ref{fig:fast_baseline_comparison} further compares the performance
profiles of Triton and CUDA generation. KernelZero achieves
43.9 fast$_1$@1 and 87 fast$_1$@10 with Triton on Level 1, compared with
17.6 and 29 with CUDA. The difference is larger on Level 2, where Triton
reaches 65.4 fast$_1$@1 and 96 fast$_1$@10, while CUDA reaches 2.4 and 12.
This gap reflects the optimization setting of the two backends. Triton
provides tile-level programming abstractions and compiler-managed lowering,
which make common tiling and fusion patterns easier to express. In contrast,
our generated CUDA kernels do not call external high-performance libraries.
Many Level-2 tasks contain convolutions or GEMMs, while their PyTorch
references directly dispatch to highly optimized cuDNN or cuBLAS kernels.
Standalone generated CUDA kernels therefore face a much stronger baseline
and rarely surpass PyTorch without library calls. The low Level-2
fast$_2$ results for both backends further show that achieving a large
speedup over these vendor-library-backed operators remains difficult.

\subsection{Ablation Study}

\paragraph{Effect of the Proposer.}
We compare KernelZero with a variant that freezes the Proposer from Step~40 and continues training the Coder on the fixed data distribution. Both settings use the same initialization and training data through Step~20. As shown in Figure~\ref{fig:proposer_training_dynamics}, continuing to update the Proposer consistently improves the final Coder. For CUDA, KernelZero improves Level-1 pass@1 from $74.9\%$ to $75.8\%$ and Level-2 pass@1 from $64.4\%$ to $69.6\%$. Its fast$_1$@1 also increases from $15.9\%$ to $17.6\%$ on Level~1 and from $2.0\%$ to $2.4\%$ on Level~2. For Triton, pass@1 improves from $74.9\%$ to $77.2\%$ on Level~1 and from $68.3\%$ to $72.5\%$ on Level~2, while fast$_1$@1 improves from $42.7\%$ to $43.9\%$ and from $61.9\%$ to $65.4\%$, respectively. The larger gains on Level~2 indicate that an updated Proposer supplies more useful training modules as the Coder becomes stronger.

\paragraph{Effect of the correctness threshold in CA-GRPO.}

We study the correctness threshold $\alpha$ in CA-GRPO while keeping all other training settings fixed. We evaluate $\alpha \in \{0, 0.5, 0.9, 1.1\}$ on the combined 200 tasks from KernelBench Levels~1 and~2. When $\alpha=0$, the speedup reward is always enabled. When $\alpha=1.1$, it is never enabled.

At Step~80, $\alpha=1.1$ achieves the highest $\mathrm{pass}@1$ of $74.85\%$, but only $9.75\%$ $\mathrm{fast}_{1.5}@1$. In comparison, $\alpha=0.9$ obtains $73.15\%$ $\mathrm{pass}@1$ and the highest $\mathrm{fast}_{1.5}@1$ of $11.00\%$. This result shows that correctness-only training limits performance optimization, while an intermediate threshold provides a better balance. Together with the per-operator results below, $\alpha=0.5$ provides the best overall configuration.

Figure~\ref{fig:reward_ablation_operators} reports the fastest correct kernel among ten samples for six representative operators. All displayed kernels were manually checked for complete output coverage and valid boundary handling. The $\alpha=0.5$ model achieves the best speedup among the four training variants on all six operators. For Tensor-MM, RL improves the SFT result from $0.12\times$ to $4.42\times$. It also reaches $1.25\times$ on GroupNorm, $1.16\times$ on Conv-BN-Scale, and $1.09\times$ on GEMM-BN-Swish. These examples show that the gated speedup reward helps the model learn effective operator-specific optimizations.

%% file: section/related.tex
\section{Related Work}

\paragraph{LLM for GPU Kernel Generation}
GPU kernel generation from high-level specifications such as PyTorch modules has attracted growing attention~\citep{kernelbench,flashinferbench}. Benchmark efforts evaluate LLM-generated kernels on correctness and runtime performance~\citep{kernelbench,tritonbench}, revealing significant gaps that motivate further training research. On the training side, supervised fine-tuning on curated kernel corpora has shown promise but requires substantial compute~\citep{cudaLLM,kernelllm,mupa}, and trained models cannot be continuously improved without repeating the full pipeline. Reinforcement learning approaches use execution feedback to optimize for correctness or speedup~\citep{drkernel,cudaagent,autotriton}, but typically treat correctness and performance as a single undifferentiated reward. Search-based and agent-based methods improve kernel quality at inference time through iterative refinement~\citep{ksearch,astra,stark}, but significantly increase inference costs.

%% file: section/conclusion.tex
\section{Conclusion}

In this work, we presented KernelZero, a framework that co-evolves a Proposer and a Coder for continuous GPU kernel generation. The Proposer generates frontier Torch modules according to the current capability of the Coder, while the Coder learns to produce correct and efficient CUDA or Triton kernels with CA-GRPO. The alternating training process forms an automatic curriculum without requiring manually constructed tasks at each stage. Experiments on KernelBench show that KernelZero-7B achieves 75.8\% and 69.6\% pass@1 for CUDA generation on Level~1 and~2, and 77.2\% and 72.5\% for Triton generation. These results demonstrate that capability-aligned data generation and correctness-aware optimization can improve both execution accuracy and kernel performance across different programming backends.

%% file: appendix/evaluate_details.tex
\section{Evaluation Details}

\subsection{Reward Server Infrastructure}

\begin{figure*}[!h]
  \begin{center}
    \centerline{\includegraphics[width=0.95\linewidth]{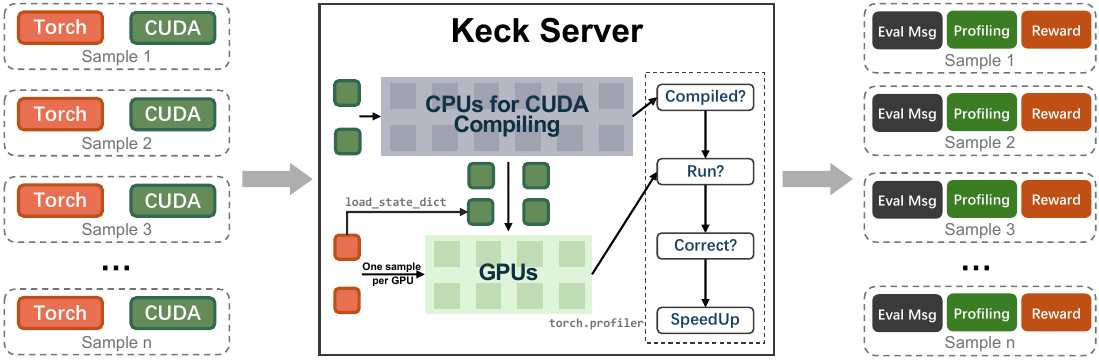}}
    \caption{
    Overview of \texttt{Keck}, the execution-based validation server for
    CUDA and Triton kernels. CUDA extensions are compiled in parallel on CPU
    workers before being dispatched to isolated GPU workers. Triton programs
    are JIT-compiled on their assigned GPUs. Both paths compare the generated
    implementation against the PyTorch reference and measure performance only
    after the correctness check passes.
    }
    \label{fig:keck}
  \end{center}
\end{figure*}

We found two issues with the original \texttt{KernelBench} evaluation.
First, running multiple programs on the same GPU can cause interference
between programs or insufficient memory, making both correctness and latency
measurements unreliable.
Second, some operators randomly initialize parameters in their
\texttt{\_\_init\_\_} methods. \texttt{KernelBench} attempts to construct
identical parameters using a fixed random seed. However, this method is
insufficient when the reference and generated implementations invoke random
operations in different orders.

To address these issues, we developed \texttt{Keck} (Kernel Check) based on
\texttt{cudaLLM}\footnote{\url{https://github.com/ByteDance-Seed/cudaLLM}}.
\texttt{Keck} provides a unified validation interface for both CUDA and Triton
implementations. It isolates candidate programs during execution and overlaps
the processing of different requests to improve evaluation throughput.
As shown in Figure~\ref{fig:keck}, \texttt{Keck} has the following features.

\begin{itemize}
    \item \textbf{Backend-aware compilation and execution.}
    CUDA extensions are compiled in parallel by CPU workers. Once compilation
    finishes, the resulting extension is immediately dispatched to an
    available GPU worker. Triton programs do not require a separate CUDA
    extension compilation stage. They are JIT-compiled when first invoked on
    the assigned GPU. The warm-up executions complete Triton compilation before
    latency is measured.

    \item \textbf{Single-sample GPU execution.}
    Each GPU worker executes only one candidate program at a time. This prevents
    memory and execution interference between samples and provides a consistent
    environment for correctness validation and latency measurement. Each
    candidate is executed in a separate process, which is terminated after the
    evaluation finishes or exceeds the time limit.

    \item \textbf{Module parameter consistency.}
    \texttt{Keck} first constructs the PyTorch reference model and the generated
    model using the same random seed. When the two implementations have
    compatible parameters, it additionally transfers the reference
    \texttt{state\_dict} to the generated CUDA or Triton model through
    \texttt{load\_state\_dict}. This ensures that correctness is evaluated using
    identical parameter values rather than relying only on identical random
    seeds. This is particularly important for operators whose initialization
    behavior depends on the construction order
    \footnote{\url{https://github.com/ScalingIntelligence/KernelBench/issues/88}}.

    \item \textbf{Unified numerical validation.}
    The reference and generated implementations receive independent copies of
    the same inputs, preventing in-place operations from affecting subsequent
    executions. Their outputs are compared using both absolute and relative
    tolerances of $10^{-2}$. Compilation errors, execution errors, timeouts,
    output-shape mismatches, and numerical mismatches are reported separately.

    \item \textbf{Backend-specific implementation constraints.}
    For CUDA generation, the target computation must be implemented using
    native CUDA kernels rather than delegated to PyTorch computational APIs.
    For Triton generation, the target computation must be performed by actual
    \texttt{@triton.jit} kernels. Python and PyTorch may be used for module
    construction and tensor allocation, but not as a replacement for the
    kernel computation being evaluated. These constraints prevent models from
    obtaining high scores by retaining the original PyTorch implementation.

    \item \textbf{Correctness-gated performance measurement.}
    Performance is measured only after the generated implementation passes the
    numerical correctness check. Both the PyTorch reference and the generated
    implementation are warmed up for two iterations and then measured for three
    iterations using CUDA events. \texttt{Keck} reports the average execution
    time and computes speedup relative to PyTorch Eager using the same inputs
    and model parameters.
\end{itemize}

%% file: appendix/grpo.tex
\section{Preliminaries: GRPO}
\label{details_grpo}
Group Relative Policy Optimization (GRPO) is an on-policy reinforcement learning algorithm built upon the Proximal Policy Optimization (PPO) framework. GRPO removes the value model to significantly reduce inference cost, while introducing group relative advantage estimation to more accurately assess the quality of model outputs. Furthermore, a KL-divergence penalty is incorporated to stabilize policy updates and prevent the policy from deviating excessively against the reference model.

Given a group $\mathcal{G}$ with rewards
$\{\mathcal{R}^{\mathcal{G}}_i\}_{i \in \mathcal{G}}$,
the group-normalized advantage is computed as
\[
\mathcal{A}^{\mathcal{G}}_i
=
\frac{
\mathcal{R}^{\mathcal{G}}_i
-
\mathrm{mean}\big(\mathcal{R}^{\mathcal{G}}_j\big)
}{
\mathrm{std}\big(\mathcal{R}^{\mathcal{G}}_j\big)
}
\]

The computed advantage is broadcast to all tokens of the corresponding output.
Model parameters are updated using the GRPO objective with a KL divergence constraint:
\[
\begin{aligned}
\mathcal{J}(\theta)
=
\mathbb{E}\Bigg[
&\frac{1}{|\mathcal{G}|}
\sum_{i \in \mathcal{G}}
\frac{1}{|o_i|}
\sum_{t=1}^{|o_i|}
\min\Big(
r_{i,t}(\theta)\mathcal{A}^{\mathcal{G}}_i,
\\
&\quad
\mathrm{clip}\big(
r_{i,t}(\theta), 1-\epsilon, 1+\epsilon
\big)
\mathcal{A}^{\mathcal{G}}_i
\Big)
\\
&\quad
-
\gamma\,
\mathrm{KL}\!\left(
\pi_{\theta}
\,\middle\|\,
\pi_{\mathrm{ref}}
\right)
\Bigg],
\\[0.5em]
r_{i,t}(\theta)
&=
\frac{
\pi_{\theta}(o_{i,t} \mid x, o_{i,<t})
}{
\pi_{\theta_{\mathrm{old}}}(o_{i,t} \mid x, o_{i,<t})
}.
\end{aligned}
\]

Here, $r_{i,t}(\theta)$ is the importance-sampling ratio at token $t$, and $\gamma$ controls the strength of the KL regularization.

%% file: appendix/hyperparams.tex
\section{Experiment Details}
\label{appendix:hyperparams}

\begin{figure}[!t]
    \centering
    \includegraphics[width=\columnwidth]{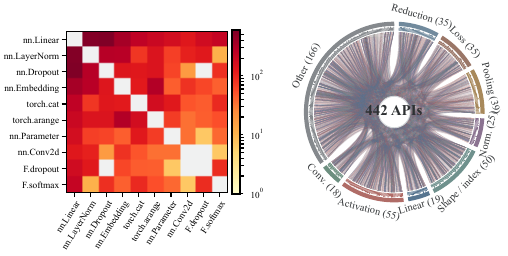}
    \caption{Torch API co-occurrence statistics and sampled API combinations
    covering all 442 candidate APIs.}
    \label{fig:api_sampling}
\end{figure}

\begin{table}[!t]
\centering
\small
\resizebox{\linewidth}{!}{
\begin{tabular}{lcc}
\toprule
\textbf{Hyperparameter} & \textbf{Proposer} & \textbf{Coder} \\
\midrule
Algorithm                         & GRPO & GRPO \\
Initialization                    & Qwen2.5-Coder-7B & KernelZero-SFT \\
RL training instances             & 400 & 1,224 \\
Policy rollout group size         & 4 & 5 \\
Nested Coder samples per module   & 5 & -- \\
Steps per round                   & 20 & 20 \\
Training rounds                   & 4 & 4 \\
Total training steps              & 80 & 80 \\
Global batch size                 & 8 & 64 \\
Rollout batch size                & 8 & 64 \\
Optimizer                         & AdamW & AdamW \\
Learning rate                     & $1.0 \times 10^{-6}$ & $1.0 \times 10^{-6}$ \\
Weight decay                      & $1.0 \times 10^{-2}$ & $1.0 \times 10^{-2}$ \\
Gradient clipping                 & 1.0 & 1.0 \\
KL coefficient ($\gamma$)         & $1.0 \times 10^{-2}$ & $1.0 \times 10^{-2}$ \\
KL penalty                        & \texttt{low\_var\_kl} & \texttt{low\_var\_kl} \\
Temperature                       & 0.6 & 0.6 \\
Top-$p$                           & 0.95 & 0.95 \\
Top-$k$                           & 20 & 20 \\
Max prompt length                 & 8,192 & 8,192 \\
Max response length               & 6,144 & 16,384 \\
Tensor parallel size              & 4 & 4 \\
Training GPUs                     & 4$\times$A100-80GB & 4$\times$A100-80GB \\
Precision                         & BF16 & BF16 \\
Approx.\ time per 20 steps        & 3.1 hours & 4.5 hours \\
\midrule
\multicolumn{3}{l}{\textit{FO-GRPO specific}} \\
Frontier target                   & 0.5 & -- \\
Invalid-module reward             & $-1$ & -- \\
\midrule
\multicolumn{3}{l}{\textit{CA-GRPO specific}} \\
Correctness threshold ($\alpha$)  & -- & 0.5 \\
Performance weight ($\beta$)      & -- & 0.1 \\
\bottomrule
\end{tabular}
}
\caption{Hyperparameters for KernelZero training. Runtime is measured for one
20-step update.}
\label{tab:hyperparams}
\end{table}

\paragraph{RL Training Data.}
The 4,000 combinations in Figure~\ref{fig:api_sampling} form the initial API-list pool. For Proposer training, we use 400 unique API-list prompts sampled from this pool. The Coder is trained on 1,224 modules that pass the validation check.

\paragraph{Training Infrastructure and Runtime.}
All optimization jobs are conducted on NVIDIA A100 GPUs with 80\,GB of memory. Each Proposer or Coder optimization job uses four A100-80GB GPUs on one node. During a Proposer update, the fixed Coder and Keck are deployed as two independent services, each using another four A100-80GB GPUs. Thus, a Proposer update occupies 12 GPUs in total, including the optimization, Coder inference, and execution-validation services. During a Coder update, Keck runs as a separate four-GPU service, resulting in eight GPUs in total.
Based on the recorded runs, a 20-step Proposer update takes approximately 3.1 hours, while a 20-step Coder update takes approximately 4.5 hours. These measurements include rollout generation, nested Coder inference, execution validation, and policy updates.

%% file: appendix/prompts.tex
\section{Prompts}

\subsection{Meta Prompt}
\label{appendix:meta_prompt}

\begin{prompt}[CUDA Coder Meta Prompt]
\textbf{Task.}
You are an expert in PyTorch and CUDA programming. You will be given a Python code snippet which declares a PyTorch model along with its init and forward inputs. The model is an instance of class \lstinline|Model|. It will be created with arguments from \lstinline|get_init_inputs()|. Then its forward function will be called with data from \lstinline|get_inputs()|. Your task is to write a custom CUDA extension to accelerate the model forward function.

\textbf{Requirements.}
\begin{itemize}
  \setlength{\itemsep}{0.2\baselineskip}
  \setlength{\parsep}{0pt}
  \setlength{\topsep}{0.25\baselineskip}
  \setlength{\partopsep}{0pt}
  \setlength{\leftmargin}{1.4em}
  \item Provide only a single Python code block in your final answer.
  \item Name your optimized model as \lstinline|ModelNew|. Keep its \lstinline|__init__| and forward function signature the same as \lstinline|Model|. Keep the names of all submodules unchanged. Ensure the keys of model state dict are unchanged. Do not create any extra tensor parameter during model initialization.
  \item Inline the CUDA code within quotes and assign it to the source variable. Inline the C++ function definition into the \lstinline|cpp_src| variable. Compile and load the extension using \lstinline|torch.utils.cpp_extension.load_inline|.
  \item Carefully decide the kernel function signature and pass the correct arguments into your kernel.
  \item Do not perform extra initialization on parameters of any submodule. Keep them initialized by default.
  \item Implement all CUDA operators by yourself. Do not call any function from the \lstinline|torch| namespace except for allocating or initializing tensors. Do not call any function from the \lstinline|torch.nn.functional| namespace. Do not call the forward function of any submodule. You can only use the parameters and attributes of the submodule. For example, pass \lstinline|self.linear.weight| and \lstinline|self.linear.bias| as arguments to your CUDA kernel instead of directly running \lstinline|self.linear(x)|.
  \item You can implement more than one kernel in the CUDA extension. If there are multiple operators within the forward function, implement all of them using as many CUDA kernels as needed.
  \item Optimize the kernel as much as possible. \textbf{DO NOT} use cuBLAS or cuDNN library calls to implement any operator.
\end{itemize}

\textbf{Target hardware: NVIDIA A100-80GB.}
\begin{itemize}
  \setlength{\itemsep}{0.1\baselineskip}
  \setlength{\parsep}{0pt}
  \setlength{\topsep}{0.2\baselineskip}
  \setlength{\partopsep}{0pt}
  \setlength{\leftmargin}{1.4em}
  \item GPU architecture: Ampere; GPU memory: 80 GB; memory bandwidth: 1935 GB/s.
  \item FP64: 9.7 TFLOPS; FP64 Tensor Core: 19.5 TFLOPS; FP32: 19.5 TFLOPS.
  \item TF32 Tensor Core: 156 TFLOPS (312 with sparsity).
  \item BFLOAT16 Tensor Core: 312 TFLOPS (624 with sparsity); FP16 Tensor Core: 312 TFLOPS (624 with sparsity).
  \item INT8 Tensor Core: 624 TOPS (1248 with sparsity).
  \item Register file: 64K 32-bit registers per SM; at most 255 registers per thread and 32 thread blocks per SM.
  \item Shared memory: 164 KB per SM and at most 163 KB per thread block.
\end{itemize}

\textbf{Optimization techniques.}
\begin{enumerate}
  \setlength{\itemsep}{0.25\baselineskip}
  \setlength{\parsep}{0pt}
  \setlength{\topsep}{0.25\baselineskip}
  \setlength{\partopsep}{0pt}
  \setlength{\leftmargin}{1.5em}
  \item \textbf{Tiling.} Partition data into tiles that fit in shared memory to reduce global memory accesses and improve memory access efficiency.

  \textit{Tiling and warp/block reduction example.}
  Given a PyTorch model: \lstinline|{example_module}|.

  \item \textbf{Fusion.} Merge consecutive operators into a single kernel to minimize memory read/write operations, especially for memory-bound workloads.

  \textit{Fusion example.}
  Given a PyTorch model: \lstinline|{example_module}|.

  \item \textbf{Pipeline.} Overlap computation with data movement to hide memory latency and maximize hardware utilization.
  \item \textbf{Reordering.} Swap loop orders to improve memory access locality, enhancing cache efficiency and reducing cache misses.
\end{enumerate}

Apply these strategies together to generate a highly efficient, memory-optimized CUDA kernel.

\textbf{Input PyTorch model.}
\begin{lstlisting}[style=kzcode,frame=none,xleftmargin=0pt,xrightmargin=0pt,linewidth=\linewidth]
__py_code_to_be_fill__
\end{lstlisting}
\end{prompt}

\subsection{Prompt of Proposer}
\label{appendix:proposer_prompt}

\begin{prompt}[Proposer Prompt]
\textbf{Task.}
You are an expert PyTorch developer. Your task is to generate a complete PyTorch \lstinline|nn.Module| that uses the following operators: \lstinline|{operators_str}|.

\textbf{Operator documentation.}

\lstinline|{docstrings_text}|

\textbf{Example format.}

Follow this exact format: \lstinline|{example_module}|.

\textbf{Requirements.}
\begin{enumerate}
  \setlength{\itemsep}{0.2\baselineskip}
  \setlength{\parsep}{0pt}
  \setlength{\topsep}{0.25\baselineskip}
  \setlength{\partopsep}{0pt}
  \setlength{\leftmargin}{1.5em}
  \item Create a class named \lstinline|Model| that inherits from \lstinline|nn.Module|.
  \item The model \textbf{MUST} use \textbf{ALL} of the following operators: \lstinline|{operators_str}|.
  \item Include proper docstrings for the class and forward method.
  \item Define \lstinline|batch_size|, input dimensions, and other necessary parameters.
  \item Implement \lstinline|get_inputs()|, which returns a list of input tensors for \lstinline|forward()|.
  \item Implement \lstinline|get_init_inputs()|, which returns a list of arguments for \lstinline|__init__()|.
  \item Use appropriate dimensions and parameters for each operator.
  \item The code should be complete and runnable.
\end{enumerate}

\textbf{Output format.}

Output \textbf{ONLY} the Python code, starting with \lstinline|import torch| and ending with the \lstinline|get_init_inputs()| function. Do \textbf{NOT} include explanations, Markdown formatting, or additional text outside the code block.

Generate the complete PyTorch module now.
\end{prompt}

%% file: appendix/case_proposer.tex
\section{Case Study on Proposer training}
\label{appendix:proposer-dag-cases}
\subsection{Qualitative Evolution of Proposer Outputs}
We compare Torch modules generated for the same API combination by the Base Proposer and three training checkpoints. The computation graphs are obtained directly from the generated modules using \texttt{torch.fx}; they are not manually reconstructed. Across these examples, training repairs operator mismatches and shape inconsistencies, and replaces mechanical operator chains with coherent branches, feature hierarchies, and task objectives.

\subsubsection{Batched multiplicative fusion}
This example exposes both kinds of multi-input structure. The Base module transforms two external inputs independently before BMM, while the final module derives two differently shaped branches from one feature tensor and merges them with BMM.

Target APIs:
\begin{itemize}
    \item \texttt{torch.nn.functional.dropout}
    \item \texttt{torch.bmm}
\end{itemize}

\ifdefined\Needspace\Needspace{169pt}\fi
\textbf{Base}\par\nopagebreak
\begin{center}
\begin{tikzpicture}[
  y=0.8cm,
  >=stealth,
  every node/.style={draw=black!65, rounded corners=1pt, fill=black!6,
    line width=0.35pt, align=center, inner sep=2.2pt,
    text width=1.85cm, minimum height=0.46cm, outer sep=0.8pt,
    font=\ttfamily\fontsize{6}{6.4}\selectfont},
  kzinput/.style={fill=white},
  kzoutput/.style={fill=black!18},
  kzparam/.style={fill=black!11},
  kzedge/.style={->, draw=black!65, line width=0.4pt,
    shorten <=1.2pt, shorten >=1.2pt}
]
\node[kzinput] (kznode0) at (-1.125,-0.000) {x\\(128,32,64)};
\node[kzinput] (kznode1) at (1.125,-0.000) {y\\(128,64,128)};
\node (kznode2) at (-1.125,-1.150) {dropout\\(128,32,64)};
\node (kznode3) at (1.125,-1.150) {dropout\\(128,64,128)};
\node (kznode4) at (0.000,-2.300) {bmm\\(128,32,128)};
\node[kzoutput] (kznode5) at (0.000,-3.450) {output\\(128,32,128)};
\draw[kzedge] (kznode0.south) -- (kznode2.north);
\draw[kzedge] (kznode1.south) -- (kznode3.north);
\draw[kzedge] (kznode2.south) -- (kznode4.north);
\draw[kzedge] (kznode3.south) -- (kznode4.north);
\draw[kzedge] (kznode4.south) -- (kznode5.north);
\end{tikzpicture}
\end{center}
\kzCodeLabel{Torch Module:}
\begin{lstlisting}[style=kzcode]
import torch
import torch.nn as nn
import torch.nn.functional as F

class Model(nn.Module):
    def __init__(self, dropout_p, batch_size, seq_len, input_dim, hidden_dim):
        super().__init__()
        self.dropout_p = dropout_p
        self.batch_size = batch_size
        self.seq_len = seq_len
        self.input_dim = input_dim
        self.hidden_dim = hidden_dim

    def forward(self, x, y):
        x = F.dropout(x, p=self.dropout_p, training=self.training)
        y = F.dropout(y, p=self.dropout_p, training=self.training)
        z = torch.bmm(x, y)
        return z

batch_size = 128
seq_len = 32
input_dim = 64
hidden_dim = 128

def get_inputs():
    x = torch.randn(batch_size, seq_len, input_dim)
    y = torch.randn(batch_size, input_dim, hidden_dim)
    return [x, y]

def get_init_inputs():
    dropout_p = 0.5
    return [dropout_p, batch_size, seq_len, input_dim, hidden_dim]
\end{lstlisting}

\ifdefined\Needspace\Needspace{254pt}\fi
\textbf{Proposer Step 20}\par\nopagebreak
\begin{center}
\begin{tikzpicture}[
  y=0.8cm,
  >=stealth,
  every node/.style={draw=black!65, rounded corners=1pt, fill=black!6,
    line width=0.35pt, align=center, inner sep=2.2pt,
    text width=1.85cm, minimum height=0.46cm, outer sep=0.8pt,
    font=\ttfamily\fontsize{6}{6.4}\selectfont},
  kzinput/.style={fill=white},
  kzoutput/.style={fill=black!18},
  kzparam/.style={fill=black!11},
  kzedge/.style={->, draw=black!65, line width=0.4pt,
    shorten <=1.2pt, shorten >=1.2pt}
]
\node[kzinput] (kznode0) at (-1.125,-0.000) {x\\(128,10)};
\node (kznode1) at (-1.125,-0.900) {unsqueeze\\(128,1,10)};
\node[kzparam] (kznode2) at (1.125,-0.000) {linear1.weight\\(20,10)};
\node (kznode3) at (1.125,-0.900) {unsqueeze\\(1,20,10)};
\node (kznode4) at (0.000,-1.800) {bmm};
\node (kznode5) at (0.000,-2.700) {squeeze};
\node (kznode6) at (-1.125,-3.600) {dropout: Dropout};
\node (kznode7) at (-1.125,-4.500) {unsqueeze};
\node[kzparam] (kznode8) at (1.125,-3.600) {linear2.weight};
\node (kznode9) at (1.125,-4.500) {unsqueeze};
\node (kznode10) at (0.000,-5.400) {bmm};
\node (kznode11) at (0.000,-6.300) {squeeze};
\node[kzoutput] (kznode12) at (0.000,-7.200) {output};
\draw[kzedge] (kznode0.south) -- (kznode1.north);
\draw[kzedge] (kznode2.south) -- (kznode3.north);
\draw[kzedge] (kznode1.south) -- (kznode4.north);
\draw[kzedge] (kznode3.south) -- (kznode4.north);
\draw[kzedge] (kznode4.south) -- (kznode5.north);
\draw[kzedge] (kznode5.south) -- (kznode6.north);
\draw[kzedge] (kznode6.south) -- (kznode7.north);
\draw[kzedge] (kznode8.south) -- (kznode9.north);
\draw[kzedge] (kznode7.south) -- (kznode10.north);
\draw[kzedge] (kznode9.south) -- (kznode10.north);
\draw[kzedge] (kznode10.south) -- (kznode11.north);
\draw[kzedge] (kznode11.south) -- (kznode12.north);
\end{tikzpicture}
\end{center}
\kzCodeLabel{Torch Module:}
\begin{lstlisting}[style=kzcode]
import torch
import torch.nn as nn

class Model(nn.Module):
    """
    Model that performs a batch matrix multiplication, applies dropout, and then another batch matrix multiplication.
    """
    def __init__(self, in_features, hidden_features, out_features, dropout_prob):
        super(Model, self).__init__()
        self.linear1 = nn.Linear(in_features, hidden_features)
        self.dropout = nn.Dropout(dropout_prob)
        self.linear2 = nn.Linear(hidden_features, out_features)

    def forward(self, x):
        """
        Args:
            x: Input tensor of shape (batch_size, in_features)
        Returns:
            Output tensor of shape (batch_size, out_features)
        """
        x = torch.bmm(x.unsqueeze(1), self.linear1.weight.unsqueeze(0)).squeeze(1)
        x = self.dropout(x)
        x = torch.bmm(x.unsqueeze(1), self.linear2.weight.unsqueeze(0)).squeeze(1)
        return x

batch_size = 128
in_features = 10
hidden_features = 20
out_features = 5
dropout_prob = 0.5

def get_inputs():
    return [torch.rand(batch_size, in_features)]

def get_init_inputs():
    return [in_features, hidden_features, out_features, dropout_prob]
\end{lstlisting}

\ifdefined\Needspace\Needspace{221pt}\fi
\textbf{Proposer Step 40}\par\nopagebreak
\begin{center}
\begin{tikzpicture}[
  y=0.8cm,
  >=stealth,
  every node/.style={draw=black!65, rounded corners=1pt, fill=black!6,
    line width=0.35pt, align=center, inner sep=2.2pt,
    text width=1.85cm, minimum height=0.46cm, outer sep=0.8pt,
    font=\ttfamily\fontsize{6}{6.4}\selectfont},
  kzinput/.style={fill=white},
  kzoutput/.style={fill=black!18},
  kzparam/.style={fill=black!11},
  kzedge/.style={->, draw=black!65, line width=0.4pt,
    shorten <=1.2pt, shorten >=1.2pt}
]
\node[kzinput] (kznode0) at (0.000,-0.000) {x\\(32,100)};
\node (kznode1) at (0.000,-1.150) {linear1: Linear\\(32,200)};
\node (kznode2) at (0.000,-2.300) {relu\\(32,200)};
\node (kznode3) at (0.000,-3.450) {dropout: Dropout\\(32,200)};
\node (kznode4) at (0.000,-4.600) {linear2: Linear\\(32,100)};
\node[kzoutput] (kznode5) at (0.000,-5.750) {output\\(32,100)};
\draw[kzedge] (kznode0.south) -- (kznode1.north);
\draw[kzedge] (kznode1.south) -- (kznode2.north);
\draw[kzedge] (kznode2.south) -- (kznode3.north);
\draw[kzedge] (kznode3.south) -- (kznode4.north);
\draw[kzedge] (kznode4.south) -- (kznode5.north);
\end{tikzpicture}
\end{center}
\kzCodeLabel{Torch Module:}
\begin{lstlisting}[style=kzcode]
import torch
import torch.nn as nn
import torch.nn.functional as F

class Model(nn.Module):
    """
    Model that performs a batch matrix multiplication, applies dropout, and then performs another batch matrix multiplication.
    """
    def __init__(self, input_dim, hidden_dim, dropout_prob):
        super(Model, self).__init__()
        self.linear1 = nn.Linear(input_dim, hidden_dim)
        self.linear2 = nn.Linear(hidden_dim, input_dim)
        self.dropout = nn.Dropout(dropout_prob)

    def forward(self, x):
        """
        Args:
            x: Input tensor of shape (batch_size, input_dim)
        Returns:
            Output tensor of shape (batch_size, input_dim)
        """
        x = self.linear1(x)
        x = F.relu(x)
        x = self.dropout(x)
        x = self.linear2(x)
        return x

batch_size = 32
input_dim = 100
hidden_dim = 200
dropout_prob = 0.5

def get_inputs():
    return [torch.randn(batch_size, input_dim)]

def get_init_inputs():
    return [input_dim, hidden_dim, dropout_prob]
\end{lstlisting}

\ifdefined\Needspace\Needspace{248pt}\fi
\textbf{Proposer Step 60}\par\nopagebreak
\begin{center}
\begin{tikzpicture}[
  y=0.8cm,
  >=stealth,
  every node/.style={draw=black!65, rounded corners=1pt, fill=black!6,
    line width=0.35pt, align=center, inner sep=2.2pt,
    text width=1.85cm, minimum height=0.46cm, outer sep=0.8pt,
    font=\ttfamily\fontsize{6}{6.4}\selectfont},
  kzinput/.style={fill=white},
  kzoutput/.style={fill=black!18},
  kzparam/.style={fill=black!11},
  kzedge/.style={->, draw=black!65, line width=0.4pt,
    shorten <=1.2pt, shorten >=1.2pt}
]
\node[kzinput] (kznode0) at (0.000,-0.000) {x\\(128,64)};
\node (kznode1) at (0.000,-1.150) {linear: Linear\\(128,32)};
\node (kznode2) at (0.000,-2.300) {dropout\\(128,32)};
\node (kznode3) at (-1.125,-3.450) {unsqueeze\\(128,1,32)};
\node (kznode4) at (1.125,-3.450) {unsqueeze\\(128,32,1)};
\node (kznode5) at (0.000,-4.600) {bmm\\(128,1,1)};
\node (kznode6) at (0.000,-5.750) {squeeze\\(128,1)};
\node[kzoutput] (kznode7) at (0.000,-6.900) {output\\(128,1)};
\draw[kzedge] (kznode0.south) -- (kznode1.north);
\draw[kzedge] (kznode1.south) -- (kznode2.north);
\draw[kzedge] (kznode2.south) -- (kznode3.north);
\draw[kzedge] (kznode2.south) -- (kznode4.north);
\draw[kzedge] (kznode3.south) -- (kznode5.north);
\draw[kzedge] (kznode4.south) -- (kznode5.north);
\draw[kzedge] (kznode5.south) -- (kznode6.north);
\draw[kzedge] (kznode6.south) -- (kznode7.north);
\end{tikzpicture}
\end{center}
\kzCodeLabel{Torch Module:}
\begin{lstlisting}[style=kzcode]
import torch
import torch.nn as nn
import torch.nn.functional as F

class Model(nn.Module):
    """
    Model that performs a batch matrix multiplication, applies dropout, and performs a forward pass.
    """
    def __init__(self, in_features, out_features):
        super(Model, self).__init__()
        self.linear = nn.Linear(in_features, out_features)

    def forward(self, x):
        """
        Args:
            x: Input tensor of shape (batch_size, in_features)
        Returns:
            Output tensor of shape (batch_size, out_features)
        """
        x = self.linear(x)
        x = F.dropout(x, p=0.5, training=self.training)
        x = torch.bmm(x.unsqueeze(1), x.unsqueeze(2)).squeeze(2)
        return x

batch_size = 128
in_features = 64
out_features = 32

def get_inputs():
    return [torch.randn(batch_size, in_features)]

def get_init_inputs():
    return [in_features, out_features]
\end{lstlisting}

\subsubsection{Multi-branch temporal convolution}
The Base module duplicates one convolution result. The final module constructs three independently parameterized convolution branches and merges their features, yielding a genuine multi-branch graph.

Target APIs:
\begin{itemize}
    \item \texttt{torch.nn.Conv1d}
    \item \texttt{torch.cat}
\end{itemize}

\ifdefined\Needspace\Needspace{169pt}\fi
\textbf{Base}\par\nopagebreak
\begin{center}
\begin{tikzpicture}[
  y=0.8cm,
  >=stealth,
  every node/.style={draw=black!65, rounded corners=1pt, fill=black!6,
    line width=0.35pt, align=center, inner sep=2.2pt,
    text width=1.85cm, minimum height=0.46cm, outer sep=0.8pt,
    font=\ttfamily\fontsize{6}{6.4}\selectfont},
  kzinput/.style={fill=white},
  kzoutput/.style={fill=black!18},
  kzparam/.style={fill=black!11},
  kzedge/.style={->, draw=black!65, line width=0.4pt,
    shorten <=1.2pt, shorten >=1.2pt}
]
\node[kzinput] (kznode0) at (0.000,-0.000) {x\\(16,3,256)};
\node (kznode1) at (0.000,-1.150) {conv1d: Conv1d\\(16,64,254)};
\node (kznode2) at (0.000,-2.300) {cat\\(16,128,254)};
\node[kzoutput] (kznode3) at (0.000,-3.450) {output\\(16,128,254)};
\draw[kzedge] (kznode0.south) -- (kznode1.north);
\draw[kzedge] (kznode1.south west) to[out=-100,in=100] (kznode2.north west);
\draw[kzedge] (kznode1.south east) to[out=-80,in=80] (kznode2.north east);
\draw[kzedge] (kznode2.south) -- (kznode3.north);
\end{tikzpicture}
\end{center}
\kzCodeLabel{Torch Module:}
\begin{lstlisting}[style=kzcode]
import torch
import torch.nn as nn

class Model(nn.Module):
    def __init__(self, in_channels, out_channels, kernel_size, stride=1, padding=0, dilation=1, bias=True):
        super().__init__()
        self.conv1d = nn.Conv1d(
            in_channels, out_channels, kernel_size, stride=stride, padding=padding, dilation=dilation, bias=bias
        )

    def forward(self, x):
        x = self.conv1d(x)
        return torch.cat([x, x], dim=1)

batch_size = 16
in_channels = 3
out_channels = 64
kernel_size = 3
width = 256

def get_inputs():
    x = torch.randn(batch_size, in_channels, width)
    return [x]

def get_init_inputs():
    return [in_channels, out_channels, kernel_size]
\end{lstlisting}

\ifdefined\Needspace\Needspace{195pt}\fi
\textbf{Proposer Step 20}\par\nopagebreak
\begin{center}
\begin{tikzpicture}[
  y=0.8cm,
  >=stealth,
  every node/.style={draw=black!65, rounded corners=1pt, fill=black!6,
    line width=0.35pt, align=center, inner sep=2.2pt,
    text width=1.85cm, minimum height=0.46cm, outer sep=0.8pt,
    font=\ttfamily\fontsize{6}{6.4}\selectfont},
  kzinput/.style={fill=white},
  kzoutput/.style={fill=black!18},
  kzparam/.style={fill=black!11},
  kzedge/.style={->, draw=black!65, line width=0.4pt,
    shorten <=1.2pt, shorten >=1.2pt}
]
\node[kzinput] (kznode0) at (0.000,-0.000) {x\\(128,3,100)};
\node (kznode1) at (0.000,-1.150) {conv: Conv1d\\(128,16,98)};
\node (kznode2) at (0.000,-2.300) {cat\\(128,32,98)};
\node (kznode3) at (0.000,-3.450) {linear: Linear};
\node[kzoutput] (kznode4) at (0.000,-4.600) {output};
\draw[kzedge] (kznode0.south) -- (kznode1.north);
\draw[kzedge] (kznode1.south west) to[out=-100,in=100] (kznode2.north west);
\draw[kzedge] (kznode1.south east) to[out=-80,in=80] (kznode2.north east);
\draw[kzedge] (kznode2.south) -- (kznode3.north);
\draw[kzedge] (kznode3.south) -- (kznode4.north);
\end{tikzpicture}
\end{center}
\kzCodeLabel{Torch Module:}
\begin{lstlisting}[style=kzcode]
import torch
import torch.nn as nn

class Model(nn.Module):
    """
    Model that performs a 1D convolution and concatenates the output with itself.
    """
    def __init__(self, in_channels, out_channels, kernel_size):
        super(Model, self).__init__()
        self.conv = nn.Conv1d(in_channels, out_channels, kernel_size)
        self.linear = nn.Linear(out_channels, out_channels)

    def forward(self, x):
        """
        Args:
            x: Input tensor of shape (batch_size, in_channels, length)
        Returns:
            Output tensor of shape (batch_size, out_channels, length)
        """
        x = self.conv(x)
        x = torch.cat((x, x), dim=1)
        x = self.linear(x)
        return x

batch_size = 128
in_channels = 3
out_channels = 16
kernel_size = 3

def get_inputs():
    return [torch.rand(batch_size, in_channels, 100)]

def get_init_inputs():
    return [in_channels, out_channels, kernel_size]
\end{lstlisting}

\ifdefined\Needspace\Needspace{169pt}\fi
\textbf{Proposer Step 40}\par\nopagebreak
\begin{center}
\begin{tikzpicture}[
  y=0.8cm,
  >=stealth,
  every node/.style={draw=black!65, rounded corners=1pt, fill=black!6,
    line width=0.35pt, align=center, inner sep=2.2pt,
    text width=1.85cm, minimum height=0.46cm, outer sep=0.8pt,
    font=\ttfamily\fontsize{6}{6.4}\selectfont},
  kzinput/.style={fill=white},
  kzoutput/.style={fill=black!18},
  kzparam/.style={fill=black!11},
  kzedge/.style={->, draw=black!65, line width=0.4pt,
    shorten <=1.2pt, shorten >=1.2pt}
]
\node[kzinput] (kznode0) at (0.000,-0.000) {x\\(128,3,100)};
\node (kznode1) at (0.000,-1.150) {conv: Conv1d\\(128,16,98)};
\node (kznode2) at (0.000,-2.300) {cat\\(128,32,98)};
\node[kzoutput] (kznode3) at (0.000,-3.450) {output\\(128,32,98)};
\draw[kzedge] (kznode0.south) -- (kznode1.north);
\draw[kzedge] (kznode1.south west) to[out=-100,in=100] (kznode2.north west);
\draw[kzedge] (kznode1.south east) to[out=-80,in=80] (kznode2.north east);
\draw[kzedge] (kznode2.south) -- (kznode3.north);
\end{tikzpicture}
\end{center}
\kzCodeLabel{Torch Module:}
\begin{lstlisting}[style=kzcode]
import torch
import torch.nn as nn

class Model(nn.Module):
    """
    Model that performs a 1D convolution and concatenates the output with itself.
    """
    def __init__(self, in_channels, out_channels, kernel_size):
        super(Model, self).__init__()
        self.conv = nn.Conv1d(in_channels, out_channels, kernel_size)
        self.cat_dim = 1

    def forward(self, x):
        """
        Args:
            x: Input tensor of shape (batch_size, in_channels, sequence_length)
        Returns:
            Output tensor of shape (batch_size, 2 * out_channels, sequence_length) where the output is concatenated with itself.
        """
        x = self.conv(x)
        x_cat = torch.cat((x, x), dim=self.cat_dim)
        return x_cat

batch_size = 128
in_channels = 3
out_channels = 16
kernel_size = 3

def get_inputs():
    return [torch.rand(batch_size, in_channels, 100)]

def get_init_inputs():
    return [in_channels, out_channels, kernel_size]
\end{lstlisting}

\ifdefined\Needspace\Needspace{169pt}\fi
\textbf{Proposer Step 60}\par\nopagebreak
\begin{center}
\begin{tikzpicture}[
  y=0.8cm,
  >=stealth,
  every node/.style={draw=black!65, rounded corners=1pt, fill=black!6,
    line width=0.35pt, align=center, inner sep=2.2pt,
    text width=1.85cm, minimum height=0.46cm, outer sep=0.8pt,
    font=\ttfamily\fontsize{6}{6.4}\selectfont},
  kzinput/.style={fill=white},
  kzoutput/.style={fill=black!18},
  kzparam/.style={fill=black!11},
  kzedge/.style={->, draw=black!65, line width=0.4pt,
    shorten <=1.2pt, shorten >=1.2pt}
]
\node[kzinput] (kznode0) at (0.000,-0.000) {x\\(128,3,100)};
\node (kznode1) at (-2.250,-1.150) {conv1: Conv1d\\(128,16,98)};
\node (kznode2) at (0.000,-1.150) {conv2: Conv1d\\(128,16,98)};
\node (kznode3) at (2.250,-1.150) {conv3: Conv1d\\(128,16,98)};
\node (kznode4) at (0.000,-2.300) {cat\\(128,48,98)};
\node[kzoutput] (kznode5) at (0.000,-3.450) {output\\(128,48,98)};
\draw[kzedge] (kznode0.south) -- (kznode1.north);
\draw[kzedge] (kznode0.south) -- (kznode2.north);
\draw[kzedge] (kznode0.south) -- (kznode3.north);
\draw[kzedge] (kznode1.south) -- (kznode4.north);
\draw[kzedge] (kznode2.south) -- (kznode4.north);
\draw[kzedge] (kznode3.south) -- (kznode4.north);
\draw[kzedge] (kznode4.south) -- (kznode5.north);
\end{tikzpicture}
\end{center}
\kzCodeLabel{Torch Module:}
\begin{lstlisting}[style=kzcode]
import torch
import torch.nn as nn

class Model(nn.Module):
    """
    Model that performs a 1D convolution and applies concatenation.
    """
    def __init__(self, in_channels, out_channels, kernel_size):
        super(Model, self).__init__()
        self.conv1 = nn.Conv1d(in_channels, out_channels, kernel_size)
        self.conv2 = nn.Conv1d(in_channels, out_channels, kernel_size)
        self.conv3 = nn.Conv1d(in_channels, out_channels, kernel_size)

    def forward(self, x):
        """
        Args:
            x: Input tensor of shape (batch_size, in_channels, length)
        Returns:
            Output tensor of shape (batch_size, out_channels, length')
        """
        x1 = self.conv1(x)
        x2 = self.conv2(x)
        x3 = self.conv3(x)
        x = torch.cat((x1, x2, x3), dim=1)
        return x

batch_size = 128
in_channels = 3
out_channels = 16
kernel_size = 3

def get_inputs():
    return [torch.rand(batch_size, in_channels, 100)]

def get_init_inputs():
    return [in_channels, out_channels, kernel_size]
\end{lstlisting}

\subsubsection{Complementary activation branches}
Training replaces repeated copies of one activation with complementary positive and negative response branches. The final module makes the two branches explicit before concatenation.

Target APIs:
\begin{itemize}
    \item \texttt{torch.nn.ReLU}
    \item \texttt{torch.cat}
\end{itemize}

\ifdefined\Needspace\Needspace{195pt}\fi
\textbf{Base}\par\nopagebreak
\begin{center}
\begin{tikzpicture}[
  y=0.8cm,
  >=stealth,
  every node/.style={draw=black!65, rounded corners=1pt, fill=black!6,
    line width=0.35pt, align=center, inner sep=2.2pt,
    text width=1.85cm, minimum height=0.46cm, outer sep=0.8pt,
    font=\ttfamily\fontsize{6}{6.4}\selectfont},
  kzinput/.style={fill=white},
  kzoutput/.style={fill=black!18},
  kzparam/.style={fill=black!11},
  kzedge/.style={->, draw=black!65, line width=0.4pt,
    shorten <=1.2pt, shorten >=1.2pt}
]
\node[kzinput] (kznode0) at (0.000,-0.000) {x\\(16,3,256,256)};
\node (kznode1) at (0.000,-1.150) {conv2d: Conv2d\\(16,64,254,252)};
\node (kznode2) at (0.000,-2.300) {relu: ReLU\\(16,64,254,252)};
\node (kznode3) at (0.000,-3.450) {cat\\(16,64,254,252)};
\node[kzoutput] (kznode4) at (0.000,-4.600) {output\\(16,64,254,252)};
\draw[kzedge] (kznode0.south) -- (kznode1.north);
\draw[kzedge] (kznode1.south) -- (kznode2.north);
\draw[kzedge] (kznode2.south) -- (kznode3.north);
\draw[kzedge] (kznode3.south) -- (kznode4.north);
\end{tikzpicture}
\end{center}
\kzCodeLabel{Torch Module:}
\begin{lstlisting}[style=kzcode]
import torch
import torch.nn as nn

class Model(nn.Module):
    def __init__(self, in_channels: int, out_channels: int, kernel_size: tuple, num_cat_dims: int):
        super().__init__()
        self.relu = nn.ReLU()
        self.cat = torch.cat
        self.conv2d = nn.Conv2d(in_channels, out_channels, kernel_size)
        self.num_cat_dims = num_cat_dims

    def forward(self, x: torch.Tensor) -> torch.Tensor:
        x = self.conv2d(x)
        x = self.relu(x)
        x = self.cat(tuple([x] * self.num_cat_dims), dim=self.num_cat_dims)
        return x

batch_size = 16
in_channels = 3
out_channels = 64
kernel_size = (3, 5)
num_cat_dims = 1
width = 256
height = 256

def get_inputs():
    x = torch.randn(batch_size, in_channels, height, width)
    return [x]

def get_init_inputs():
    return [in_channels, out_channels, kernel_size, num_cat_dims]
\end{lstlisting}

\ifdefined\Needspace\Needspace{248pt}\fi
\textbf{Proposer Step 20}\par\nopagebreak
\begin{center}
\begin{tikzpicture}[
  y=0.8cm,
  >=stealth,
  every node/.style={draw=black!65, rounded corners=1pt, fill=black!6,
    line width=0.35pt, align=center, inner sep=2.2pt,
    text width=1.85cm, minimum height=0.46cm, outer sep=0.8pt,
    font=\ttfamily\fontsize{6}{6.4}\selectfont},
  kzinput/.style={fill=white},
  kzoutput/.style={fill=black!18},
  kzparam/.style={fill=black!11},
  kzedge/.style={->, draw=black!65, line width=0.4pt,
    shorten <=1.2pt, shorten >=1.2pt}
]
\node[kzinput] (kznode0) at (0.000,-0.000) {x\\(128,3,32,32)};
\node (kznode1) at (0.000,-1.150) {conv: Conv2d\\(128,16,32,32)};
\node (kznode2) at (0.000,-2.300) {relu: ReLU\\(128,16,32,32)};
\node (kznode3) at (0.000,-3.450) {neg\\(128,16,32,32)};
\node (kznode4) at (0.000,-4.600) {relu: ReLU\\(128,16,32,32)};
\node (kznode5) at (0.000,-5.750) {cat\\(128,32,32,32)};
\node[kzoutput] (kznode6) at (0.000,-6.900) {output\\(128,32,32,32)};
\draw[kzedge] (kznode0.south) -- (kznode1.north);
\draw[kzedge] (kznode1.south) -- (kznode2.north);
\draw[kzedge] (kznode2.south) -- (kznode3.north);
\draw[kzedge] (kznode3.south) -- (kznode4.north);
\draw[kzedge] (kznode2.south) -- (kznode5.north);
\draw[kzedge] (kznode4.south) -- (kznode5.north);
\draw[kzedge] (kznode5.south) -- (kznode6.north);
\end{tikzpicture}
\end{center}
\kzCodeLabel{Torch Module:}
\begin{lstlisting}[style=kzcode]
import torch
import torch.nn as nn

class Model(nn.Module):
    """
    Model that applies ReLU and concatenates the output with its negative.
    """
    def __init__(self, in_channels, out_channels):
        super(Model, self).__init__()
        self.conv = nn.Conv2d(in_channels, out_channels, kernel_size=3, padding=1)
        self.relu = nn.ReLU()

    def forward(self, x):
        """
        Args:
            x: Input tensor of shape (batch_size, in_channels, height, width)
        Returns:
            Output tensor of shape (batch_size, out_channels*2, height, width)
        """
        x = self.conv(x)
        x = self.relu(x)
        neg_x = self.relu(-x)
        x = torch.cat((x, neg_x), dim=1)
        return x

batch_size = 128
in_channels = 3
out_channels = 16

def get_inputs():
    return [torch.rand(batch_size, in_channels, 32, 32)]

def get_init_inputs():
    return [in_channels, out_channels]
\end{lstlisting}

\ifdefined\Needspace\Needspace{248pt}\fi
\textbf{Proposer Step 40}\par\nopagebreak
\begin{center}
\begin{tikzpicture}[
  y=0.8cm,
  >=stealth,
  every node/.style={draw=black!65, rounded corners=1pt, fill=black!6,
    line width=0.35pt, align=center, inner sep=2.2pt,
    text width=1.85cm, minimum height=0.46cm, outer sep=0.8pt,
    font=\ttfamily\fontsize{6}{6.4}\selectfont},
  kzinput/.style={fill=white},
  kzoutput/.style={fill=black!18},
  kzparam/.style={fill=black!11},
  kzedge/.style={->, draw=black!65, line width=0.4pt,
    shorten <=1.2pt, shorten >=1.2pt}
]
\node[kzinput] (kznode0) at (0.000,-0.000) {x\\(128,3,32,32)};
\node (kznode1) at (0.000,-1.150) {conv: Conv2d\\(128,16,32,32)};
\node (kznode2) at (0.000,-2.300) {relu: ReLU\\(128,16,32,32)};
\node (kznode3) at (0.000,-3.450) {neg\\(128,16,32,32)};
\node (kznode4) at (0.000,-4.600) {relu: ReLU\\(128,16,32,32)};
\node (kznode5) at (0.000,-5.750) {cat\\(128,32,32,32)};
\node[kzoutput] (kznode6) at (0.000,-6.900) {output\\(128,32,32,32)};
\draw[kzedge] (kznode0.south) -- (kznode1.north);
\draw[kzedge] (kznode1.south) -- (kznode2.north);
\draw[kzedge] (kznode2.south) -- (kznode3.north);
\draw[kzedge] (kznode3.south) -- (kznode4.north);
\draw[kzedge] (kznode2.south) -- (kznode5.north);
\draw[kzedge] (kznode4.south) -- (kznode5.north);
\draw[kzedge] (kznode5.south) -- (kznode6.north);
\end{tikzpicture}
\end{center}
\kzCodeLabel{Torch Module:}
\begin{lstlisting}[style=kzcode]
import torch
import torch.nn as nn

class Model(nn.Module):
    """
    Model that applies ReLU and concatenates the output with its negative.
    """
    def __init__(self, in_channels, out_channels):
        super(Model, self).__init__()
        self.conv = nn.Conv2d(in_channels, out_channels, kernel_size=3, padding=1)
        self.relu = nn.ReLU()

    def forward(self, x):
        """
        Args:
            x: Input tensor of shape (batch_size, in_channels, height, width)
        Returns:
            Output tensor of shape (batch_size, out_channels*2, height, width)
        """
        x = self.conv(x)
        x = self.relu(x)
        neg_x = self.relu(-x)
        x = torch.cat((x, neg_x), dim=1)
        return x

batch_size = 128
in_channels = 3
out_channels = 16

def get_inputs():
    return [torch.rand(batch_size, in_channels, 32, 32)]

def get_init_inputs():
    return [in_channels, out_channels]
\end{lstlisting}

\ifdefined\Needspace\Needspace{221pt}\fi
\textbf{Proposer Step 60}\par\nopagebreak
\begin{center}
\begin{tikzpicture}[
  y=0.8cm,
  >=stealth,
  every node/.style={draw=black!65, rounded corners=1pt, fill=black!6,
    line width=0.35pt, align=center, inner sep=2.2pt,
    text width=1.85cm, minimum height=0.46cm, outer sep=0.8pt,
    font=\ttfamily\fontsize{6}{6.4}\selectfont},
  kzinput/.style={fill=white},
  kzoutput/.style={fill=black!18},
  kzparam/.style={fill=black!11},
  kzedge/.style={->, draw=black!65, line width=0.4pt,
    shorten <=1.2pt, shorten >=1.2pt}
]
\node[kzinput] (kznode0) at (0.000,-0.000) {x\\(128,3,32,32)};
\node (kznode1) at (0.000,-1.150) {conv: Conv2d\\(128,16,32,32)};
\node (kznode2) at (-1.125,-2.300) {relu1: ReLU\\(128,16,32,32)};
\node (kznode3) at (1.125,-2.300) {neg\\(128,16,32,32)};
\node (kznode4) at (0.000,-3.450) {relu2: ReLU\\(128,16,32,32)};
\node (kznode5) at (0.000,-4.600) {cat\\(128,32,32,32)};
\node[kzoutput] (kznode6) at (0.000,-5.750) {output\\(128,32,32,32)};
\draw[kzedge] (kznode0.south) -- (kznode1.north);
\draw[kzedge] (kznode1.south) -- (kznode2.north);
\draw[kzedge] (kznode1.south) -- (kznode3.north);
\draw[kzedge] (kznode3.south) -- (kznode4.north);
\draw[kzedge] (kznode2.south) -- (kznode5.north);
\draw[kzedge] (kznode4.south) -- (kznode5.north);
\draw[kzedge] (kznode5.south) -- (kznode6.north);
\end{tikzpicture}
\end{center}
\kzCodeLabel{Torch Module:}
\begin{lstlisting}[style=kzcode]
import torch
import torch.nn as nn

class Model(nn.Module):
    """
    Model that applies ReLU and concatenates outputs from two ReLU layers.
    """
    def __init__(self, in_channels, out_channels):
        super(Model, self).__init__()
        self.conv = nn.Conv2d(in_channels, out_channels, kernel_size=3, padding=1)
        self.relu1 = nn.ReLU()
        self.relu2 = nn.ReLU()

    def forward(self, x):
        """
        Args:
            x: Input tensor of shape (batch_size, in_channels, height, width)
        Returns:
            Output tensor of shape (batch_size, out_channels * 2, height', width') where height', width' are the dimensions after convolution.
        """
        x = self.conv(x)
        x1 = self.relu1(x)
        x2 = self.relu2(-x)
        x = torch.cat((x1, x2), dim=1)
        return x

batch_size = 128
in_channels = 3
out_channels = 16

def get_inputs():
    return [torch.rand(batch_size, in_channels, 32, 32)]

def get_init_inputs():
    return [in_channels, out_channels]
\end{lstlisting}

\subsubsection{Embedding classifier with an explicit objective}
The module contains an explicit objective merge. The feature branch passes through embedding and a two-layer classifier, while the target enters the loss node through a separate external-input edge.

Target APIs:
\begin{itemize}
    \item \texttt{torch.cat}
    \item \texttt{torch.nn.CrossEntropyLoss}
    \item \texttt{torch.nn.Embedding}
\end{itemize}

\ifdefined\Needspace\Needspace{195pt}\fi
\textbf{Base}\par\nopagebreak
\begin{center}
\begin{tikzpicture}[
  y=0.8cm,
  >=stealth,
  every node/.style={draw=black!65, rounded corners=1pt, fill=black!6,
    line width=0.35pt, align=center, inner sep=2.2pt,
    text width=1.85cm, minimum height=0.46cm, outer sep=0.8pt,
    font=\ttfamily\fontsize{6}{6.4}\selectfont},
  kzinput/.style={fill=white},
  kzoutput/.style={fill=black!18},
  kzparam/.style={fill=black!11},
  kzedge/.style={->, draw=black!65, line width=0.4pt,
    shorten <=1.2pt, shorten >=1.2pt}
]
\node[kzinput] (kznode0) at (-2.250,-0.000) {x1\\(64,)};
\node[kzinput] (kznode1) at (0.000,-0.000) {x2\\(64,)};
\node[kzinput] (kznode2) at (2.250,-0.000) {target\\(64,)};
\node (kznode3) at (-1.125,-1.150) {embedding:\\Embedding};
\node (kznode4) at (1.125,-1.150) {embedding:\\Embedding};
\node (kznode5) at (0.000,-2.300) {cat\\(64,1024)};
\node (kznode6) at (0.000,-3.450) {CrossEntropyLoss\\()};
\node[kzoutput] (kznode7) at (0.000,-4.600) {output\\()};
\draw[kzedge] (kznode0.south) -- (kznode3.north);
\draw[kzedge] (kznode1.south) -- (kznode4.north);
\draw[kzedge] (kznode3.south) -- (kznode5.north);
\draw[kzedge] (kznode4.south) -- (kznode5.north);
\draw[kzedge] (kznode5.south) -- (kznode6.north);
\draw[kzedge] (kznode2.south) -- (kznode6.north);
\draw[kzedge] (kznode6.south) -- (kznode7.north);
\end{tikzpicture}
\end{center}
\kzCodeLabel{Torch Module:}
\begin{lstlisting}[style=kzcode]
import torch
import torch.nn as nn

class Model(nn.Module):
    def __init__(self, num_embeddings, embedding_dim, num_classes):
        super().__init__()
        self.embedding = nn.Embedding(num_embeddings, embedding_dim)
        self.cross_entropy_loss = nn.CrossEntropyLoss()

    def forward(self, x1, x2, target):
        emb1 = self.embedding(x1)
        emb2 = self.embedding(x2)
        cat_emb = torch.cat((emb1, emb2), dim=1)
        loss = self.cross_entropy_loss(cat_emb, target)
        return loss

num_embeddings = 10
embedding_dim = 512
num_classes = 10
batch_size = 64

def get_inputs():
    x1 = torch.randint(0, num_embeddings, (batch_size,))
    x2 = torch.randint(0, num_embeddings, (batch_size,))
    target = torch.randint(0, num_classes, (batch_size,))
    return [x1, x2, target]

def get_init_inputs():
    return [num_embeddings, embedding_dim, num_classes]
\end{lstlisting}

\ifdefined\Needspace\Needspace{195pt}\fi
\textbf{Proposer Step 20}\par\nopagebreak
\begin{center}
\begin{tikzpicture}[
  y=0.8cm,
  >=stealth,
  every node/.style={draw=black!65, rounded corners=1pt, fill=black!6,
    line width=0.35pt, align=center, inner sep=2.2pt,
    text width=1.85cm, minimum height=0.46cm, outer sep=0.8pt,
    font=\ttfamily\fontsize{6}{6.4}\selectfont},
  kzinput/.style={fill=white},
  kzoutput/.style={fill=black!18},
  kzparam/.style={fill=black!11},
  kzedge/.style={->, draw=black!65, line width=0.4pt,
    shorten <=1.2pt, shorten >=1.2pt}
]
\node[kzinput] (kznode0) at (0.000,-0.000) {x\\(32,10)};
\node (kznode1) at (0.000,-1.150) {embedding:\\Embedding};
\node (kznode2) at (0.000,-2.300) {mean\\(32,128)};
\node (kznode3) at (0.000,-3.450) {linear: Linear\\(32,10)};
\node[kzoutput] (kznode4) at (0.000,-4.600) {output\\(32,10)};
\draw[kzedge] (kznode0.south) -- (kznode1.north);
\draw[kzedge] (kznode1.south) -- (kznode2.north);
\draw[kzedge] (kznode2.south) -- (kznode3.north);
\draw[kzedge] (kznode3.south) -- (kznode4.north);
\end{tikzpicture}
\end{center}
\kzCodeLabel{Torch Module:}
\begin{lstlisting}[style=kzcode]
import torch
import torch.nn as nn

class Model(nn.Module):
    """
    Model that uses torch.cat, torch.nn.CrossEntropyLoss, and torch.nn.Embedding.
    """
    def __init__(self, vocab_size, embedding_dim, num_classes):
        super(Model, self).__init__()
        self.embedding = nn.Embedding(vocab_size, embedding_dim)
        self.linear = nn.Linear(embedding_dim, num_classes)

    def forward(self, x):
        """
        Args:
            x: Input tensor of shape (batch_size, sequence_length)
        Returns:
            Output tensor of shape (batch_size, num_classes)
        """
        x = self.embedding(x)
        x = torch.mean(x, dim=1)
        x = self.linear(x)
        return x

batch_size = 32
vocab_size = 1000
embedding_dim = 128
num_classes = 10

def get_inputs():
    return [torch.randint(0, vocab_size, (batch_size, 10))]

def get_init_inputs():
    return [vocab_size, embedding_dim, num_classes]
\end{lstlisting}

\ifdefined\Needspace\Needspace{248pt}\fi
\textbf{Proposer Step 40}\par\nopagebreak
\begin{center}
\begin{tikzpicture}[
  y=0.8cm,
  >=stealth,
  every node/.style={draw=black!65, rounded corners=1pt, fill=black!6,
    line width=0.35pt, align=center, inner sep=2.2pt,
    text width=1.85cm, minimum height=0.46cm, outer sep=0.8pt,
    font=\ttfamily\fontsize{6}{6.4}\selectfont},
  kzinput/.style={fill=white},
  kzoutput/.style={fill=black!18},
  kzparam/.style={fill=black!11},
  kzedge/.style={->, draw=black!65, line width=0.4pt,
    shorten <=1.2pt, shorten >=1.2pt}
]
\node[kzinput] (kznode0) at (0.000,-0.000) {x\\(64,10)};
\node (kznode1) at (0.000,-1.150) {embedding:\\Embedding};
\node (kznode2) at (0.000,-2.300) {cat\\(64,20,128)};
\node (kznode3) at (0.000,-3.450) {fc1: Linear\\(64,20,256)};
\node (kznode4) at (0.000,-4.600) {relu\\(64,20,256)};
\node (kznode5) at (0.000,-5.750) {fc2: Linear\\(64,20,5)};
\node[kzoutput] (kznode6) at (0.000,-6.900) {output\\(64,20,5)};
\draw[kzedge] (kznode0.south) -- (kznode1.north);
\draw[kzedge] (kznode1.south west) to[out=-100,in=100] (kznode2.north west);
\draw[kzedge] (kznode1.south east) to[out=-80,in=80] (kznode2.north east);
\draw[kzedge] (kznode2.south) -- (kznode3.north);
\draw[kzedge] (kznode3.south) -- (kznode4.north);
\draw[kzedge] (kznode4.south) -- (kznode5.north);
\draw[kzedge] (kznode5.south) -- (kznode6.north);
\end{tikzpicture}
\end{center}
\kzCodeLabel{Torch Module:}
\begin{lstlisting}[style=kzcode]
import torch
import torch.nn as nn

class Model(nn.Module):
    """
    Model that uses torch.cat, torch.nn.CrossEntropyLoss, and torch.nn.Embedding.
    """
    def __init__(self, num_embeddings, embedding_dim, input_size, hidden_size, output_size):
        super(Model, self).__init__()
        self.embedding = nn.Embedding(num_embeddings, embedding_dim)
        self.fc1 = nn.Linear(embedding_dim, hidden_size)
        self.fc2 = nn.Linear(hidden_size, output_size)

    def forward(self, x):
        """
        Args:
            x: Input tensor of shape (batch_size, input_size)
        Returns:
            Output tensor of shape (batch_size, output_size)
        """
        x = self.embedding(x)
        x = torch.cat((x, x), dim=1)
        x = torch.relu(self.fc1(x))
        x = self.fc2(x)
        return x

batch_size = 64
num_embeddings = 1000
embedding_dim = 128
input_size = 10
hidden_size = 256
output_size = 5

def get_inputs():
    return [torch.randint(0, num_embeddings, (batch_size, input_size))]

def get_init_inputs():
    return [num_embeddings, embedding_dim, input_size, hidden_size, output_size]
\end{lstlisting}

\ifdefined\Needspace\Needspace{254pt}\fi
\textbf{Proposer Step 60}\par\nopagebreak
\begin{center}
\begin{tikzpicture}[
  y=0.8cm,
  >=stealth,
  every node/.style={draw=black!65, rounded corners=1pt, fill=black!6,
    line width=0.35pt, align=center, inner sep=2.2pt,
    text width=1.85cm, minimum height=0.46cm, outer sep=0.8pt,
    font=\ttfamily\fontsize{6}{6.4}\selectfont},
  kzinput/.style={fill=white},
  kzoutput/.style={fill=black!18},
  kzparam/.style={fill=black!11},
  kzedge/.style={->, draw=black!65, line width=0.4pt,
    shorten <=1.2pt, shorten >=1.2pt}
]
\node[kzinput] (kznode0) at (-1.125,-0.000) {x\\(128,)};
\node[kzinput] (kznode1) at (1.125,-0.000) {y\\(128,)};
\node (kznode2) at (0.000,-1.029) {embedding:\\Embedding};
\node (kznode3) at (0.000,-2.057) {cat\\(128,16)};
\node (kznode4) at (0.000,-3.086) {fc1: Linear\\(128,64)};
\node (kznode5) at (0.000,-4.114) {relu\\(128,64)};
\node (kznode6) at (0.000,-5.143) {fc2: Linear\\(128,10)};
\node (kznode7) at (0.000,-6.171) {CrossEntropyLoss\\()};
\node[kzoutput] (kznode8) at (0.000,-7.200) {output\\()};
\draw[kzedge] (kznode0.south) -- (kznode2.north);
\draw[kzedge] (kznode2.south west) to[out=-100,in=100] (kznode3.north west);
\draw[kzedge] (kznode2.south east) to[out=-80,in=80] (kznode3.north east);
\draw[kzedge] (kznode3.south) -- (kznode4.north);
\draw[kzedge] (kznode4.south) -- (kznode5.north);
\draw[kzedge] (kznode5.south) -- (kznode6.north);
\draw[kzedge] (kznode6.south) -- (kznode7.north);
\draw[kzedge] (kznode1.south) -- (kznode7.north);
\draw[kzedge] (kznode7.south) -- (kznode8.north);
\end{tikzpicture}
\end{center}
\kzCodeLabel{Torch Module:}
\begin{lstlisting}[style=kzcode]
import torch
import torch.nn as nn

class Model(nn.Module):
    """
    Model that uses torch.cat, torch.nn.CrossEntropyLoss, and torch.nn.Embedding.
    """
    def __init__(self, num_embeddings, embedding_dim):
        super(Model, self).__init__()
        self.embedding = torch.nn.Embedding(num_embeddings, embedding_dim)
        self.fc1 = nn.Linear(embedding_dim * 2, 64)
        self.fc2 = nn.Linear(64, num_embeddings)
        self.loss_fn = torch.nn.CrossEntropyLoss()

    def forward(self, x, y):
        """
        Args:
            x: Input tensor of shape (batch_size, num_embeddings)
            y: Target tensor of shape (batch_size)
        Returns:
            Loss tensor
        """
        x = self.embedding(x)
        x = torch.cat((x, x), dim=1)
        x = torch.relu(self.fc1(x))
        x = self.fc2(x)
        loss = self.loss_fn(x, y)
        return loss

batch_size = 128
num_embeddings = 10
embedding_dim = 8

def get_inputs():
    x = torch.randint(0, num_embeddings, (batch_size,))
    y = torch.randint(0, num_embeddings, (batch_size,))
    return [x, y]

def get_init_inputs():
    return [num_embeddings, embedding_dim]
\end{lstlisting}

\subsubsection{Parallel volumetric pooling fusion}
The Base module omits the requested concatenation. The final module shares a convolutional feature tensor across two distinct pooling branches and concatenates their outputs.

Target APIs:
\begin{itemize}
    \item \texttt{torch.nn.Dropout}
    \item \texttt{torch.cat}
\end{itemize}

\ifdefined\Needspace\Needspace{291pt}\fi
\textbf{Base}\par\nopagebreak
\begin{center}
\begin{tikzpicture}[
  y=0.8cm,
  >=stealth,
  every node/.style={draw=black!65, rounded corners=1pt, fill=black!6,
    line width=0.35pt, align=center, inner sep=2.2pt,
    text width=1.85cm, minimum height=0.46cm, outer sep=0.8pt,
    font=\ttfamily\fontsize{6}{6.4}\selectfont},
  kzinput/.style={fill=white},
  kzoutput/.style={fill=black!18},
  kzparam/.style={fill=black!11},
  kzedge/.style={->, draw=black!65, line width=0.4pt,
    shorten <=1.2pt, shorten >=1.2pt}
]
\node[kzinput] (kznode0) at (0.000,-0.000) {x\\(128,3,32,32)};
\node (kznode1) at (0.000,-0.880) {conv: Conv2d\\(128,16,30,30)};
\node (kznode2) at (0.000,-1.760) {relu\\(128,16,30,30)};
\node (kznode3) at (0.000,-2.640) {dropout: Dropout\\(128,16,30,30)};
\node (kznode4) at (0.000,-3.520) {layers.0: Conv2d\\(128,16,28,28)};
\node (kznode5) at (0.000,-4.400) {relu\\(128,16,28,28)};
\node (kznode6) at (0.000,-5.280) {dropout: Dropout\\(128,16,28,28)};
\node (kznode7) at (0.000,-6.160) {layers.1: Conv2d\\(128,16,26,26)};
\node (kznode8) at (0.000,-7.040) {relu\\(128,16,26,26)};
\node (kznode9) at (0.000,-7.920) {dropout: Dropout\\(128,16,26,26)};
\node[kzoutput] (kznode10) at (0.000,-8.800) {output\\(128,16,26,26)};
\draw[kzedge] (kznode0.south) -- (kznode1.north);
\draw[kzedge] (kznode1.south) -- (kznode2.north);
\draw[kzedge] (kznode2.south) -- (kznode3.north);
\draw[kzedge] (kznode3.south) -- (kznode4.north);
\draw[kzedge] (kznode4.south) -- (kznode5.north);
\draw[kzedge] (kznode5.south) -- (kznode6.north);
\draw[kzedge] (kznode6.south) -- (kznode7.north);
\draw[kzedge] (kznode7.south) -- (kznode8.north);
\draw[kzedge] (kznode8.south) -- (kznode9.north);
\draw[kzedge] (kznode9.south) -- (kznode10.north);
\end{tikzpicture}
\end{center}
\kzCodeLabel{Torch Module:}
\begin{lstlisting}[style=kzcode]
import torch
import torch.nn as nn

class Model(nn.Module):
    def __init__(self, in_channels, out_channels, kernel_size, dropout_prob, num_layers):
        super().__init__()
        self.conv = nn.Conv2d(in_channels, out_channels, kernel_size)
        self.dropout = nn.Dropout(p=dropout_prob)
        self.layers = nn.ModuleList([nn.Conv2d(out_channels, out_channels, kernel_size) for _ in range(num_layers - 1)])

    def forward(self, x):
        x = self.conv(x)
        x = torch.relu(x)
        x = self.dropout(x)
        for layer in self.layers:
            x = layer(x)
            x = torch.relu(x)
            x = self.dropout(x)
        return x

batch_size = 128
in_channels = 3
out_channels = 16
height, width = 32, 32
kernel_size = 3
dropout_prob = 0.5
num_layers = 3

def get_inputs():
    return [torch.randn(batch_size, in_channels, height, width)]

def get_init_inputs():
    return [in_channels, out_channels, kernel_size, dropout_prob, num_layers]
\end{lstlisting}

\ifdefined\Needspace\Needspace{221pt}\fi
\textbf{Proposer Step 20}\par\nopagebreak
\begin{center}
\begin{tikzpicture}[
  y=0.8cm,
  >=stealth,
  every node/.style={draw=black!65, rounded corners=1pt, fill=black!6,
    line width=0.35pt, align=center, inner sep=2.2pt,
    text width=1.85cm, minimum height=0.46cm, outer sep=0.8pt,
    font=\ttfamily\fontsize{6}{6.4}\selectfont},
  kzinput/.style={fill=white},
  kzoutput/.style={fill=black!18},
  kzparam/.style={fill=black!11},
  kzedge/.style={->, draw=black!65, line width=0.4pt,
    shorten <=1.2pt, shorten >=1.2pt}
]
\node[kzinput] (kznode0) at (0.000,-0.000) {x\\(128,3,16,32,32)};
\node (kznode1) at (0.000,-1.150) {conv: Conv3d\\(128,1...30,30)};
\node (kznode2) at (0.000,-2.300) {dropout: Dropout\\(128,1...30,30)};
\node (kznode3) at (0.000,-3.450) {cat\\(128,3...30,30)};
\node (kznode4) at (0.000,-4.600) {pool: MaxPool3d\\(128,32,7,15,15)};
\node[kzoutput] (kznode5) at (0.000,-5.750) {output\\(128,32,7,15,15)};
\draw[kzedge] (kznode0.south) -- (kznode1.north);
\draw[kzedge] (kznode1.south) -- (kznode2.north);
\draw[kzedge] (kznode2.south west) to[out=-100,in=100] (kznode3.north west);
\draw[kzedge] (kznode2.south east) to[out=-80,in=80] (kznode3.north east);
\draw[kzedge] (kznode3.south) -- (kznode4.north);
\draw[kzedge] (kznode4.south) -- (kznode5.north);
\end{tikzpicture}
\end{center}
\kzCodeLabel{Torch Module:}
\begin{lstlisting}[style=kzcode]
import torch
import torch.nn as nn

class Model(nn.Module):
    """
    Model that performs a 3D convolution, applies Dropout, concatenates the output with itself, and performs a max pooling operation.
    """
    def __init__(self, in_channels, out_channels, kernel_size, dropout_prob):
        super(Model, self).__init__()
        self.conv = nn.Conv3d(in_channels, out_channels, kernel_size)
        self.dropout = nn.Dropout(dropout_prob)
        self.pool = nn.MaxPool3d(2)

    def forward(self, x):
        """
        Args:
            x: Input tensor of shape (batch_size, in_channels, depth, height, width)
        Returns:
            Output tensor of shape (batch_size, 2*out_channels, depth', height', width') where depth', height', width' are the dimensions after pooling.
        """
        x = self.conv(x)
        x = self.dropout(x)
        x = torch.cat((x, x), dim=1)
        x = self.pool(x)
        return x

batch_size = 128
in_channels = 3
out_channels = 16
kernel_size = 3
dropout_prob = 0.2

def get_inputs():
    return [torch.rand(batch_size, in_channels, 16, 32, 32)]

def get_init_inputs():
    return [in_channels, out_channels, kernel_size, dropout_prob]
\end{lstlisting}

\ifdefined\Needspace\Needspace{221pt}\fi
\textbf{Proposer Step 40}\par\nopagebreak
\begin{center}
\begin{tikzpicture}[
  y=0.8cm,
  >=stealth,
  every node/.style={draw=black!65, rounded corners=1pt, fill=black!6,
    line width=0.35pt, align=center, inner sep=2.2pt,
    text width=1.85cm, minimum height=0.46cm, outer sep=0.8pt,
    font=\ttfamily\fontsize{6}{6.4}\selectfont},
  kzinput/.style={fill=white},
  kzoutput/.style={fill=black!18},
  kzparam/.style={fill=black!11},
  kzedge/.style={->, draw=black!65, line width=0.4pt,
    shorten <=1.2pt, shorten >=1.2pt}
]
\node[kzinput] (kznode0) at (0.000,-0.000) {x\\(128,3,16,32,32)};
\node (kznode1) at (0.000,-1.150) {conv: Conv3d\\(128,1...30,30)};
\node (kznode2) at (0.000,-2.300) {dropout: Dropout\\(128,1...30,30)};
\node (kznode3) at (-1.125,-3.450) {pool1: MaxPool3d\\(128,16,7,15,15)};
\node (kznode4) at (1.125,-3.450) {pool2: MaxPool3d\\(128,16,7,15,15)};
\node (kznode5) at (0.000,-4.600) {cat\\(128,32,7,15,15)};
\node[kzoutput] (kznode6) at (0.000,-5.750) {output\\(128,32,7,15,15)};
\draw[kzedge] (kznode0.south) -- (kznode1.north);
\draw[kzedge] (kznode1.south) -- (kznode2.north);
\draw[kzedge] (kznode2.south) -- (kznode3.north);
\draw[kzedge] (kznode2.south) -- (kznode4.north);
\draw[kzedge] (kznode3.south) -- (kznode5.north);
\draw[kzedge] (kznode4.south) -- (kznode5.north);
\draw[kzedge] (kznode5.south) -- (kznode6.north);
\end{tikzpicture}
\end{center}
\kzCodeLabel{Torch Module:}
\begin{lstlisting}[style=kzcode]
import torch
import torch.nn as nn

class Model(nn.Module):
    """
    Model that performs a 3D convolution, applies Dropout, concatenates the output, and performs two max pooling operations.
    """
    def __init__(self, in_channels, out_channels, kernel_size, dropout_prob, pool_kernel_size):
        super(Model, self).__init__()
        self.conv = nn.Conv3d(in_channels, out_channels, kernel_size)
        self.dropout = nn.Dropout(dropout_prob)
        self.pool1 = nn.MaxPool3d(pool_kernel_size)
        self.pool2 = nn.MaxPool3d(pool_kernel_size)

    def forward(self, x):
        """
        Args:
            x: Input tensor of shape (batch_size, in_channels, depth, height, width)
        Returns:
            Output tensor of shape (batch_size, out_channels, depth', height', width') where depth', height', width' are the dimensions after pooling.
        """
        x = self.conv(x)
        x = self.dropout(x)
        x1 = self.pool1(x)
        x2 = self.pool2(x)
        x = torch.cat((x1, x2), dim=1)
        return x

batch_size = 128
in_channels = 3
out_channels = 16
kernel_size = 3
dropout_prob = 0.5
pool_kernel_size = 2

def get_inputs():
    return [torch.rand(batch_size, in_channels, 16, 32, 32)]

def get_init_inputs():
    return [in_channels, out_channels, kernel_size, dropout_prob, pool_kernel_size]
\end{lstlisting}

\ifdefined\Needspace\Needspace{221pt}\fi
\textbf{Proposer Step 60}\par\nopagebreak
\begin{center}
\begin{tikzpicture}[
  y=0.8cm,
  >=stealth,
  every node/.style={draw=black!65, rounded corners=1pt, fill=black!6,
    line width=0.35pt, align=center, inner sep=2.2pt,
    text width=1.85cm, minimum height=0.46cm, outer sep=0.8pt,
    font=\ttfamily\fontsize{6}{6.4}\selectfont},
  kzinput/.style={fill=white},
  kzoutput/.style={fill=black!18},
  kzparam/.style={fill=black!11},
  kzedge/.style={->, draw=black!65, line width=0.4pt,
    shorten <=1.2pt, shorten >=1.2pt}
]
\node[kzinput] (kznode0) at (0.000,-0.000) {x\\(128,3,16,32,32)};
\node (kznode1) at (0.000,-1.150) {conv: Conv3d\\(128,1...30,30)};
\node (kznode2) at (0.000,-2.300) {dropout: Dropout\\(128,1...30,30)};
\node (kznode3) at (-1.125,-3.450) {pool1: MaxPool3d\\(128,16,7,15,15)};
\node (kznode4) at (1.125,-3.450) {pool2: MaxPool3d\\(128,16,7,15,15)};
\node (kznode5) at (0.000,-4.600) {cat\\(128,32,7,15,15)};
\node[kzoutput] (kznode6) at (0.000,-5.750) {output\\(128,32,7,15,15)};
\draw[kzedge] (kznode0.south) -- (kznode1.north);
\draw[kzedge] (kznode1.south) -- (kznode2.north);
\draw[kzedge] (kznode2.south) -- (kznode3.north);
\draw[kzedge] (kznode2.south) -- (kznode4.north);
\draw[kzedge] (kznode3.south) -- (kznode5.north);
\draw[kzedge] (kznode4.south) -- (kznode5.north);
\draw[kzedge] (kznode5.south) -- (kznode6.north);
\end{tikzpicture}
\end{center}
\kzCodeLabel{Torch Module:}
\begin{lstlisting}[style=kzcode]
import torch
import torch.nn as nn

class Model(nn.Module):
    """
    Model that performs a 3D convolution, applies Dropout, concatenates the output, and performs two max pooling operations.
    """
    def __init__(self, in_channels, out_channels, kernel_size, dropout_prob, pool_kernel_size):
        super(Model, self).__init__()
        self.conv = nn.Conv3d(in_channels, out_channels, kernel_size)
        self.dropout = nn.Dropout(dropout_prob)
        self.pool1 = nn.MaxPool3d(pool_kernel_size)
        self.pool2 = nn.MaxPool3d(pool_kernel_size)

    def forward(self, x):
        """
        Args:
            x: Input tensor of shape (batch_size, in_channels, depth, height, width)
        Returns:
            Output tensor of shape (batch_size, out_channels, depth', height', width') where depth', height', width' are the dimensions after pooling.
        """
        x = self.conv(x)
        x = self.dropout(x)
        x1 = self.pool1(x)
        x2 = self.pool2(x)
        x = torch.cat((x1, x2), dim=1)
        return x

batch_size = 128
in_channels = 3
out_channels = 16
kernel_size = 3
dropout_prob = 0.5
pool_kernel_size = 2

def get_inputs():
    return [torch.rand(batch_size, in_channels, 16, 32, 32)]

def get_init_inputs():
    return [in_channels, out_channels, kernel_size, dropout_prob, pool_kernel_size]
\end{lstlisting}

\subsubsection{Shared trunk with two output heads}
The final module builds a shared volumetric feature trunk and then splits it into a global mean head and an elementwise square-root head. Both outputs are returned explicitly.

Target APIs:
\begin{itemize}
    \item \texttt{torch.sqrt}
    \item \texttt{torch.mean}
\end{itemize}

\ifdefined\Needspace\Needspace{169pt}\fi
\textbf{Base}\par\nopagebreak
\begin{center}
\begin{tikzpicture}[
  y=0.8cm,
  >=stealth,
  every node/.style={draw=black!65, rounded corners=1pt, fill=black!6,
    line width=0.35pt, align=center, inner sep=2.2pt,
    text width=1.85cm, minimum height=0.46cm, outer sep=0.8pt,
    font=\ttfamily\fontsize{6}{6.4}\selectfont},
  kzinput/.style={fill=white},
  kzoutput/.style={fill=black!18},
  kzparam/.style={fill=black!11},
  kzedge/.style={->, draw=black!65, line width=0.4pt,
    shorten <=1.2pt, shorten >=1.2pt}
]
\node[kzinput] (kznode0) at (0.000,-0.000) {x\\(128,3,32,32)};
\node (kznode1) at (0.000,-1.150) {sqrt\\(128,3,32,32)};
\node (kznode2) at (0.000,-2.300) {mean\\()};
\node[kzoutput] (kznode3) at (0.000,-3.450) {output\\()};
\draw[kzedge] (kznode0.south) -- (kznode1.north);
\draw[kzedge] (kznode1.south) -- (kznode2.north);
\draw[kzedge] (kznode2.south) -- (kznode3.north);
\end{tikzpicture}
\end{center}
\kzCodeLabel{Torch Module:}
\begin{lstlisting}[style=kzcode]
import torch
import torch.nn as nn
import torch.nn.functional as F

class Model(nn.Module):
    def __init__(self):
        super().__init__()

    def forward(self, x):
        x = torch.sqrt(x)
        x = torch.mean(x)
        return x

batch_size = 128
channels = 3
height, width = 32, 32

def get_inputs():
    return [torch.randn(batch_size, channels, height, width).abs()]  # Ensure input is non-negative for sqrt

def get_init_inputs():
    return []
\end{lstlisting}

\ifdefined\Needspace\Needspace{221pt}\fi
\textbf{Proposer Step 20}\par\nopagebreak
\begin{center}
\begin{tikzpicture}[
  y=0.8cm,
  >=stealth,
  every node/.style={draw=black!65, rounded corners=1pt, fill=black!6,
    line width=0.35pt, align=center, inner sep=2.2pt,
    text width=1.85cm, minimum height=0.46cm, outer sep=0.8pt,
    font=\ttfamily\fontsize{6}{6.4}\selectfont},
  kzinput/.style={fill=white},
  kzoutput/.style={fill=black!18},
  kzparam/.style={fill=black!11},
  kzedge/.style={->, draw=black!65, line width=0.4pt,
    shorten <=1.2pt, shorten >=1.2pt}
]
\node[kzinput] (kznode0) at (0.000,-0.000) {x\\(128,3,16,32,32)};
\node (kznode1) at (0.000,-1.150) {conv: Conv3d\\(128,1...30,30)};
\node (kznode2) at (0.000,-2.300) {softmax\\(128,1...30,30)};
\node (kznode3) at (0.000,-3.450) {mean\\()};
\node (kznode4) at (0.000,-4.600) {sqrt\\()};
\node[kzoutput] (kznode5) at (0.000,-5.750) {output\\()};
\draw[kzedge] (kznode0.south) -- (kznode1.north);
\draw[kzedge] (kznode1.south) -- (kznode2.north);
\draw[kzedge] (kznode2.south) -- (kznode3.north);
\draw[kzedge] (kznode3.south) -- (kznode4.north);
\draw[kzedge] (kznode4.south) -- (kznode5.north);
\end{tikzpicture}
\end{center}
\kzCodeLabel{Torch Module:}
\begin{lstlisting}[style=kzcode]
import torch
import torch.nn as nn

class Model(nn.Module):
    """
    Model that performs a 3D convolution, applies Softmax, and computes the mean and square root of the output.
    """
    def __init__(self, in_channels, out_channels, kernel_size):
        super(Model, self).__init__()
        self.conv = nn.Conv3d(in_channels, out_channels, kernel_size)

    def forward(self, x):
        """
        Args:
            x: Input tensor of shape (batch_size, in_channels, depth, height, width)
        Returns:
            Output tensor of shape (batch_size, out_channels, depth', height', width') where depth', height', width' are the dimensions after convolution.
        """
        x = self.conv(x)
        x = torch.softmax(x, dim=1)
        x_mean = torch.mean(x)
        x_sqrt = torch.sqrt(x_mean)
        return x_sqrt

batch_size = 128
in_channels = 3
out_channels = 16
kernel_size = 3

def get_inputs():
    return [torch.rand(batch_size, in_channels, 16, 32, 32)]

def get_init_inputs():
    return [in_channels, out_channels, kernel_size]
\end{lstlisting}

\ifdefined\Needspace\Needspace{195pt}\fi
\textbf{Proposer Step 40}\par\nopagebreak
\begin{center}
\begin{tikzpicture}[
  y=0.8cm,
  >=stealth,
  every node/.style={draw=black!65, rounded corners=1pt, fill=black!6,
    line width=0.35pt, align=center, inner sep=2.2pt,
    text width=1.85cm, minimum height=0.46cm, outer sep=0.8pt,
    font=\ttfamily\fontsize{6}{6.4}\selectfont},
  kzinput/.style={fill=white},
  kzoutput/.style={fill=black!18},
  kzparam/.style={fill=black!11},
  kzedge/.style={->, draw=black!65, line width=0.4pt,
    shorten <=1.2pt, shorten >=1.2pt}
]
\node[kzinput] (kznode0) at (0.000,-0.000) {x\\(128,100)};
\node (kznode1) at (0.000,-1.150) {sqrt\\(128,100)};
\node (kznode2) at (0.000,-2.300) {mean\\(128,)};
\node (kznode3) at (0.000,-3.450) {linear: Linear};
\node[kzoutput] (kznode4) at (0.000,-4.600) {output};
\draw[kzedge] (kznode0.south) -- (kznode1.north);
\draw[kzedge] (kznode1.south) -- (kznode2.north);
\draw[kzedge] (kznode2.south) -- (kznode3.north);
\draw[kzedge] (kznode3.south) -- (kznode4.north);
\end{tikzpicture}
\end{center}
\kzCodeLabel{Torch Module:}
\begin{lstlisting}[style=kzcode]
import torch
import torch.nn as nn

class Model(nn.Module):
    """
    Model that performs a square root operation, computes the mean, and applies a linear transformation.
    """
    def __init__(self, in_features, out_features):
        super(Model, self).__init__()
        self.linear = nn.Linear(in_features, out_features)

    def forward(self, x):
        """
        Args:
            x: Input tensor of shape (batch_size, in_features)
        Returns:
            Output tensor of shape (batch_size, out_features)
        """
        x = torch.sqrt(x)
        x = torch.mean(x, dim=1)
        x = self.linear(x)
        return x

batch_size = 128
in_features = 100
out_features = 50

def get_inputs():
    return [torch.rand(batch_size, in_features)]

def get_init_inputs():
    return [in_features, out_features]
\end{lstlisting}

\ifdefined\Needspace\Needspace{248pt}\fi
\textbf{Proposer Step 60}\par\nopagebreak
\begin{center}
\begin{tikzpicture}[
  y=0.8cm,
  >=stealth,
  every node/.style={draw=black!65, rounded corners=1pt, fill=black!6,
    line width=0.35pt, align=center, inner sep=2.2pt,
    text width=1.85cm, minimum height=0.46cm, outer sep=0.8pt,
    font=\ttfamily\fontsize{6}{6.4}\selectfont},
  kzinput/.style={fill=white},
  kzoutput/.style={fill=black!18},
  kzparam/.style={fill=black!11},
  kzedge/.style={->, draw=black!65, line width=0.4pt,
    shorten <=1.2pt, shorten >=1.2pt}
]
\node[kzinput] (kznode0) at (0.000,-0.000) {x\\(128,3,16,32,32)};
\node (kznode1) at (0.000,-1.150) {conv: Conv3d\\(128,1...30,30)};
\node (kznode2) at (0.000,-2.300) {softmax\\(128,1...30,30)};
\node (kznode3) at (0.000,-3.450) {pool1: MaxPool3d\\(128,16,7,15,15)};
\node (kznode4) at (0.000,-4.600) {pool2: MaxPool3d\\(128,16,3,7,7)};
\node (kznode5) at (-1.125,-5.750) {mean\\()};
\node (kznode6) at (1.125,-5.750) {sqrt\\(128,16,3,7,7)};
\node[kzoutput] (kznode7) at (0.000,-6.900) {output\\(),(12...3,7,7)};
\draw[kzedge] (kznode0.south) -- (kznode1.north);
\draw[kzedge] (kznode1.south) -- (kznode2.north);
\draw[kzedge] (kznode2.south) -- (kznode3.north);
\draw[kzedge] (kznode3.south) -- (kznode4.north);
\draw[kzedge] (kznode4.south) -- (kznode5.north);
\draw[kzedge] (kznode4.south) -- (kznode6.north);
\draw[kzedge] (kznode5.south) -- (kznode7.north);
\draw[kzedge] (kznode6.south) -- (kznode7.north);
\end{tikzpicture}
\end{center}
\kzCodeLabel{Torch Module:}
\begin{lstlisting}[style=kzcode]
import torch
import torch.nn as nn

class Model(nn.Module):
    """
    Model that performs a 3D convolution, applies Softmax, and calculates the mean and square root of the output.
    """
    def __init__(self, in_channels, out_channels, kernel_size, pool_kernel_size):
        super(Model, self).__init__()
        self.conv = nn.Conv3d(in_channels, out_channels, kernel_size)
        self.pool1 = nn.MaxPool3d(pool_kernel_size)
        self.pool2 = nn.MaxPool3d(pool_kernel_size)

    def forward(self, x):
        """
        Args:
            x: Input tensor of shape (batch_size, in_channels, depth, height, width)
        Returns:
            Output tensor of shape (batch_size, out_channels, depth', height', width') where depth', height', width' are the dimensions after pooling.
        """
        x = self.conv(x)
        x = torch.softmax(x, dim=1)
        x = self.pool1(x)
        x = self.pool2(x)
        mean_value = torch.mean(x)
        sqrt_value = torch.sqrt(x)
        return mean_value, sqrt_value

batch_size = 128
in_channels = 3
out_channels = 16
depth, height, width = 16, 32, 32
kernel_size = 3
pool_kernel_size = 2

def get_inputs():
    return [torch.rand(batch_size, in_channels, depth, height, width)]

def get_init_inputs():
    return [in_channels, out_channels, kernel_size, pool_kernel_size]
\end{lstlisting}

%% file: appendix/case_backend.tex
\section{Case Study on Generated Kernels}
The following examples are selected from evaluation records by jointly considering reported speedup and implementation complexity. The CUDA examples come from the verified CUDA evaluation runs. Reported speedups are relative to PyTorch Eager on an NVIDIA A100-80GB GPU.

\subsection{CUDA Examples}

\subsubsection{Diagonal Matrix Multiplication}
\noindent\textbf{KernelBench task:} 1-12\quad\textbf{Measured speedup:} $13.59\times$
\kzCodeLabel{Torch Module:}
\begin{lstlisting}[style=kzcode]
import torch
import torch.nn as nn

class Model(nn.Module):
    """
    Simple model that performs a matrix multiplication of a diagonal matrix with another matrix.
    C = diag(A) * B
    """
    def __init__(self):
        super(Model, self).__init__()
    
    def forward(self, A, B):
        """
        Performs the matrix multiplication.

        Args:
            A (torch.Tensor): A 1D tensor representing the diagonal of the diagonal matrix. Shape: (N,).
            B (torch.Tensor): A 2D tensor representing the second matrix. Shape: (N, M).

        Returns:
            torch.Tensor: The result of the matrix multiplication. Shape: (N, M).
        """
        return torch.diag(A) @ B

M = 4096
N = 4096

def get_inputs():
    A = torch.rand(N)
    B = torch.rand(N, M)
    return [A, B]

def get_init_inputs():
    return []  # No special initialization inputs needed
\end{lstlisting}
\kzCodeLabel{CUDA generated by KernelZero:}
\begin{lstlisting}[style=kzcode]
import torch
import torch.nn as nn
from torch.utils.cpp_extension import load_inline

# ----------------------------------------------------------------------
# CUDA kernel: C[i, j] = A[i] * B[i, j]
# Uses tiling (32x32) and a small shared-memory tile for the vector A.
# ----------------------------------------------------------------------
cuda_source = r'''
#include <torch/extension.h>
#include <ATen/cuda/CUDAContext.h>
#include <cuda_runtime.h>

#ifndef TILE_N
#define TILE_N 32          // rows per block
#endif
#ifndef TILE_M
#define TILE_M 32          // columns per block
#endif

// ------------------------------------------------------------------
// Each block tiles a TILE_N x TILE_M region of the output matrix.
// The slice of A that corresponds to the rows handled by the block
// is loaded once into shared memory (one float per row).
// ------------------------------------------------------------------
__global__ void diag_mul_kernel(const float* __restrict__ A,
                                const float* __restrict__ B,
                                float* __restrict__ C,
                                int N,
                                int M) {
    // Global indices
    int row = blockIdx.x * TILE_N + threadIdx.x;   // row index
    int col = blockIdx.y * TILE_M + threadIdx.y;   // column index

    // Shared memory: one element of A for each row in the tile
    __shared__ float As[TILE_N];

    // Load A[row] into shared memory - only one thread per row does it
    if (threadIdx.y == 0 && row < N) {
        As[threadIdx.x] = A[row];
    }
    __syncthreads();

    // Compute output if we are inside the valid matrix bounds
    if (row < N && col < M) {
        float a = As[threadIdx.x];                 // A[row] (cached)
        float b = B[row * M + col];                // B[row, col]
        C[row * M + col] = a * b;                  // C[row, col]
    }
}

// ------------------------------------------------------------------
// C++ wrapper called from Python
// ------------------------------------------------------------------
torch::Tensor diag_mul_cuda(torch::Tensor A, torch::Tensor B) {
    TORCH_CHECK(A.is_cuda() && B.is_cuda(),
                "Both inputs must be CUDA tensors");
    TORCH_CHECK(A.dtype() == torch::kFloat32 && B.dtype() == torch::kFloat32,
                "Only float32 tensors are supported");
    TORCH_CHECK(A.dim() == 1 && B.dim() == 2,
                "A must be 1-D and B must be 2-D");
    TORCH_CHECK(A.size(0) == B.size(0),
                "The first dimension of B must match the size of A");

    const int N = A.size(0);
    const int M = B.size(1);

    // Allocate output tensor
    auto C = torch::empty({N, M}, A.options());

    // Launch configuration
    dim3 blockDim(TILE_N, TILE_M);
    dim3 gridDim((N + TILE_N - 1) / TILE_N,
                 (M + TILE_M - 1) / TILE_M);

    // Use the current CUDA stream
    cudaStream_t stream = at::cuda::getCurrentCUDAStream();

    diag_mul_kernel<<<gridDim, blockDim, 0, stream>>>(
        A.data_ptr<float>(),            // compatible with older PyTorch versions
        B.data_ptr<float>(),
        C.data_ptr<float>(),
        N,
        M
    );

    return C;
}
'''

# ----------------------------------------------------------------------
# Minimal C++ source needed for compilation (includes the proper header)
# ----------------------------------------------------------------------
cpp_source = r'''
#include <torch/extension.h>

torch::Tensor diag_mul_cuda(torch::Tensor A, torch::Tensor B);
'''

# Compile the inline CUDA extension
diag_mul_module = load_inline(
    name="diag_mul_ext",
    cpp_sources=cpp_source,
    cuda_sources=cuda_source,
    functions=["diag_mul_cuda"],
    verbose=False,
    extra_cflags=["-O3"],
    extra_cuda_cflags=["-O3", "--use_fast_math"],
)

# ----------------------------------------------------------------------
# Optimized model that delegates the computation to the custom kernel
# ----------------------------------------------------------------------
class ModelNew(nn.Module):
    def __init__(self):
        super().__init__()   # identical signature to the original Model

    def forward(self, A, B):
        """
        Args:
            A (torch.Tensor): 1-D tensor of shape (N,)
            B (torch.Tensor): 2-D tensor of shape (N, M)
        Returns:
            torch.Tensor: (N, M) = diag(A) @ B
        """
        return diag_mul_module.diag_mul_cuda(A, B)

# ----------------------------------------------------------------------
# Optional simple test (can be removed in production)
# ----------------------------------------------------------------------
if __name__ == "__main__":
    N, M = 4096, 4096
    A = torch.rand(N, device="cuda")
    B = torch.rand(N, M, device="cuda")
    model = ModelNew().cuda()
    C = model(A, B)
    # Verify correctness (tolerances may be loosened for large tensors)
    # ref = torch.diag(A) @ B
    # assert torch.allclose(C, ref, atol=1e-5, rtol=1e-4)
\end{lstlisting}

\subsubsection{Cross-Entropy Loss}
\noindent\textbf{KernelBench task:} 1-95\quad\textbf{Measured speedup:} $4.79\times$
\kzCodeLabel{Torch Module:}
\begin{lstlisting}[style=kzcode]
import torch
import torch.nn as nn

class Model(nn.Module):
    """
    A model that computes Cross Entropy Loss for multi-class classification tasks.

    Parameters:
        None
    """
    def __init__(self):
        super(Model, self).__init__()

    def forward(self, predictions, targets):
        return torch.nn.functional.cross_entropy(predictions, targets)

batch_size = 32768
num_classes = 4096
input_shape = (num_classes,)
dim = 1

def get_inputs():
    return [torch.rand(batch_size, *input_shape), torch.randint(0, num_classes, (batch_size,))]

def get_init_inputs():
    return []
\end{lstlisting}
\kzCodeLabel{CUDA generated by KernelZero:}
\begin{lstlisting}[style=kzcode]
import torch
import torch.nn as nn
from torch.utils.cpp_extension import load_inline

# -------------------------------------------------
# CUDA source implementing a numerically-stable
# cross-entropy (log-softmax + negative log-likelihood)
# with mean reduction.  One block processes one sample.
# -------------------------------------------------
cuda_source = r'''
#include <torch/extension.h>
#include <cuda_runtime.h>
#include <math.h>

#ifndef WARP_SIZE
#define WARP_SIZE 32
#endif

// ----- block-level max reduction (no extra helpers) -----
__device__ float block_max_reduce(float val) {
    // warp reduction
    #pragma unroll
    for (int offset = WARP_SIZE / 2; offset > 0; offset /= 2) {
        val = fmaxf(val, __shfl_xor_sync(0xffffffff, val, offset));
    }
    // write warp leaders to shared memory
    __shared__ float warp_max[WARP_SIZE];          // enough for up to 32 warps
    int lane = threadIdx.x % WARP_SIZE;
    int wid  = threadIdx.x / WARP_SIZE;
    if (lane == 0) warp_max[wid] = val;
    __syncthreads();

    // second stage: reduce the warp leaders
    float block_max = (threadIdx.x < blockDim.x / WARP_SIZE) ? warp_max[threadIdx.x] : -INFINITY;
    if (wid == 0) {   // only the first warp does the final reduction
        #pragma unroll
        for (int offset = WARP_SIZE / 2; offset > 0; offset /= 2) {
            block_max = fmaxf(block_max, __shfl_xor_sync(0xffffffff, block_max, offset));
        }
    }
    return block_max;
}

// ----- block-level sum reduction (similar) -----
__device__ float block_sum_reduce(float val) {
    #pragma unroll
    for (int offset = WARP_SIZE / 2; offset > 0; offset /= 2) {
        val += __shfl_xor_sync(0xffffffff, val, offset);
    }
    __shared__ float warp_sum[WARP_SIZE];
    int lane = threadIdx.x % WARP_SIZE;
    int wid  = threadIdx.x / WARP_SIZE;
    if (lane == 0) warp_sum[wid] = val;
    __syncthreads();

    float block_sum = (threadIdx.x < blockDim.x / WARP_SIZE) ? warp_sum[threadIdx.x] : 0.f;
    if (wid == 0) {
        #pragma unroll
        for (int offset = WARP_SIZE / 2; offset > 0; offset /= 2) {
            block_sum += __shfl_xor_sync(0xffffffff, block_sum, offset);
        }
    }
    return block_sum;
}

// ------------------- cross-entropy kernel -------------------
extern "C"
__global__ void cross_entropy_kernel(const float* __restrict__ logits,
                                     const int64_t* __restrict__ targets,
                                     float* __restrict__ total_loss,
                                     const int N,
                                     const int C) {
    int sample = blockIdx.x;                 // one block per sample
    if (sample >= N) return;

    const int tid   = threadIdx.x;
    const int stride = blockDim.x;
    const float* row = logits + (size_t)sample * C;

    // ---- 1) max over classes (for numerical stability) ----
    float max_val = -INFINITY;
    for (int c = tid; c < C; c += stride) {
        max_val = fmaxf(max_val, row[c]);
    }
    max_val = block_max_reduce(max_val);
    __shared__ float s_max;
    if (tid == 0) s_max = max_val;
    __syncthreads();
    max_val = s_max;

    // ---- 2) sum of exp(logits - max) ----
    float sum_exp = 0.f;
    for (int c = tid; c < C; c += stride) {
        sum_exp += expf(row[c] - max_val);
    }
    sum_exp = block_sum_reduce(sum_exp);
    __shared__ float s_sum;
    if (tid == 0) s_sum = sum_exp;
    __syncthreads();
    sum_exp = s_sum;

    // ---- 3) per-sample loss ----
    float log_sum_exp = max_val + logf(sum_exp);
    int64_t target = targets[sample];
    float logit_t = row[target];
    float loss = -(logit_t - log_sum_exp);   // -log softmax

    // ---- 4) accumulate into global total (single thread per block) ----
    if (tid == 0) {
        atomicAdd(total_loss, loss);
    }
}

// ------------------- C++ wrapper -------------------
torch::Tensor cross_entropy_cuda(torch::Tensor logits, torch::Tensor targets) {
    TORCH_CHECK(logits.is_cuda(), "logits must be a CUDA tensor");
    TORCH_CHECK(targets.is_cuda(), "targets must be a CUDA tensor");
    TORCH_CHECK(logits.dim() == 2, "logits must be 2-D (N, C)");
    TORCH_CHECK(targets.dim() == 1, "targets must be 1-D (N)");
    TORCH_CHECK(logits.dtype() == torch::kFloat32, "logits dtype must be float32");
    TORCH_CHECK(targets.dtype() == torch::kInt64,  "targets dtype must be int64");

    const int N = logits.size(0);
    const int C = logits.size(1);

    // scalar to hold the accumulated loss
    auto total = torch::zeros({}, logits.options());

    const int threads = 256;
    const dim3 blocks(N);

    cross_entropy_kernel<<<blocks, threads>>>(
        logits.data_ptr<float>(),
        targets.data_ptr<int64_t>(),
        total.data_ptr<float>(),
        N,
        C
    );

    // check for launch errors
    cudaError_t err = cudaGetLastError();
    if (err != cudaSuccess) {
        AT_ERROR("CUDA error in cross_entropy_kernel: ", cudaGetErrorString(err));
    }

    // make sure kernel finished before we read the result
    torch::cuda::synchronize();

    // mean reduction
    return total / static_cast<float>(N);
}
'''

# -------------------------------------------------
# C++ header (declaration) - include necessary headers
# -------------------------------------------------
cpp_source = r'''
#include <torch/extension.h>
torch::Tensor cross_entropy_cuda(torch::Tensor logits, torch::Tensor targets);
'''

# Compile the inline extension
cross_entropy_ext = load_inline(
    name="cross_entropy_ext",
    cpp_sources=cpp_source,
    cuda_sources=cuda_source,
    functions=["cross_entropy_cuda"],
    verbose=False,
    extra_cflags=["-O3"],
    extra_cuda_cflags=["-O3", "--use_fast_math"],
)

# -------------------------------------------------
# Optimized model using the custom kernel
# -------------------------------------------------
class ModelNew(nn.Module):
    def __init__(self):
        super().__init__()

    def forward(self, predictions, targets):
        """
        predictions : Tensor of shape (batch, num_classes), dtype float32, CUDA
        targets     : Tensor of shape (batch,),      dtype int64,   CUDA
        Returns a 0-dim tensor containing the mean cross-entropy loss.
        """
        return cross_entropy_ext.cross_entropy_cuda(predictions, targets)
\end{lstlisting}

\subsubsection{Triplet-Margin Loss}
\noindent\textbf{KernelBench task:} 1-99\quad\textbf{Measured speedup:} $3.88\times$
\kzCodeLabel{Torch Module:}
\begin{lstlisting}[style=kzcode]
import torch
import torch.nn as nn

class Model(nn.Module):
    """
    A model that computes Triplet Margin Loss for metric learning tasks.

    Parameters:
        margin (float): The margin between the positive and negative samples.
    """
    def __init__(self, margin=1.0):
        super(Model, self).__init__()
        self.loss_fn = torch.nn.TripletMarginLoss(margin=margin)

    def forward(self, anchor, positive, negative):
        return self.loss_fn(anchor, positive, negative)

batch_size = 32768
input_shape = (8192,)
dim = 1

def get_inputs():
    scale = torch.rand(())
    return [torch.rand(batch_size, *input_shape)*scale, torch.rand(batch_size, *input_shape), torch.rand(batch_size, *input_shape)]
    
def get_init_inputs():
    return [1.0]  # Default margin
\end{lstlisting}
\kzCodeLabel{CUDA generated by KernelZero:}
\begin{lstlisting}[style=kzcode]
import torch
import torch.nn as nn
from torch.utils.cpp_extension import load_inline

# ------------------------------------------------------------------
# CUDA kernel implementing TripletMarginLoss (mean reduction, p=2,
# eps=1e-6, no swap). One block processes one sample; within the block
# threads accumulate squared distances and reduce them. A second
# kernel finishes the mean reduction.
# ------------------------------------------------------------------
cuda_source = r"""
#include <torch/extension.h>
#include <cuda_runtime.h>
#include <ATen/cuda/CUDAContext.h>
#include <math.h>

#ifndef WARP_SIZE
#define WARP_SIZE 32
#endif

// ---------- warp and block reduction helpers ----------
__device__ __forceinline__ float warp_reduce_sum(float val) {
#pragma unroll
    for (int offset = WARP_SIZE/2; offset > 0; offset /= 2)
        val += __shfl_down_sync(0xffffffff, val, offset);
    return val;
}

__device__ __forceinline__ float block_reduce_sum(float val) {
    static __shared__ float shared[WARP_SIZE];   // up to 32 warps per block
    int lane = threadIdx.x % WARP_SIZE;
    int wid  = threadIdx.x / WARP_SIZE;

    val = warp_reduce_sum(val);                // warp-level reduction
    if (lane == 0) shared[wid] = val;          // store per-warp result
    __syncthreads();

    // final reduction by first warp
    val = (threadIdx.x < blockDim.x / WARP_SIZE) ? shared[lane] : 0.0f;
    if (wid == 0) val = warp_reduce_sum(val);
    return val;
}

// -------------------- main loss kernel --------------------
extern "C" __global__
void triplet_margin_loss_kernel(const float* __restrict__ anchor,
                                const float* __restrict__ positive,
                                const float* __restrict__ negative,
                                float* __restrict__ loss_sum,
                                const int dim,
                                const float margin) {
    const int sample_idx = blockIdx.x;          // one block per batch element
    const float eps = 1e-6f;

    const float* a = anchor   + sample_idx * dim;
    const float* p = positive + sample_idx * dim;
    const float* n = negative + sample_idx * dim;

    float sq_ap = 0.0f;
    float sq_an = 0.0f;

    // each thread works on a strided slice of the feature dimension
    for (int i = threadIdx.x; i < dim; i += blockDim.x) {
        float diff_ap = a[i] - p[i];
        float diff_an = a[i] - n[i];
        sq_ap += diff_ap * diff_ap;
        sq_an += diff_an * diff_an;
    }

    // block-wide reduction to get full squared-norms
    float sum_ap = block_reduce_sum(sq_ap);
    float sum_an = block_reduce_sum(sq_an);

    if (threadIdx.x == 0) {
        float dist_ap = sqrtf(sum_ap + eps);
        float dist_an = sqrtf(sum_an + eps);
        float loss_i  = fmaxf(0.0f, dist_ap - dist_an + margin);
        atomicAdd(loss_sum, loss_i);            // accumulate across batch
    }
}

// ------------- final division kernel (mean reduction) -------------
extern "C" __global__
void divide_by_batch_kernel(float* __restrict__ loss_sum, const int batch) {
    if (threadIdx.x == 0 && blockIdx.x == 0) {
        loss_sum[0] = loss_sum[0] / static_cast<float>(batch);
    }
}

// ------------------- C++ wrapper -------------------
torch::Tensor triplet_margin_loss_cuda(torch::Tensor anchor,
                                      torch::Tensor positive,
                                      torch::Tensor negative,
                                      const float margin) {
    // sanity checks - they are also performed on the Python side,
    // but keeping them here makes the kernel more robust.
    TORCH_CHECK(anchor.is_cuda() && positive.is_cuda() && negative.is_cuda(),
                "All inputs must be CUDA tensors");
    TORCH_CHECK(anchor.device() == positive.device() &&
                anchor.device() == negative.device(),
                "All inputs must reside on the same device");

    const int batch = anchor.size(0);
    const int dim   = anchor.size(1);

    // scalar tensor that will hold the summed loss - initialized to zero
    auto loss = torch::zeros({}, anchor.options());

    const int threads = 256;
    const int blocks  = batch;   // one block per sample

    // launch main kernel
    triplet_margin_loss_kernel<<<blocks, threads>>>(
        anchor.data_ptr<float>(),
        positive.data_ptr<float>(),
        negative.data_ptr<float>(),
        loss.data_ptr<float>(),
        dim,
        margin
    );

    // launch kernel that divides by batch size to obtain the mean
    divide_by_batch_kernel<<<1, 1>>>(loss.data_ptr<float>(), batch);

    // optional: synchronize and check for launch errors (helps debugging)
    cudaError_t err = cudaGetLastError();
    if (err != cudaSuccess) {
        AT_ERROR("CUDA kernel launch failed: ", cudaGetErrorString(err));
    }

    return loss;   // scalar (0-dim) tensor
}
"""

cpp_source = r"""
torch::Tensor triplet_margin_loss_cuda(torch::Tensor anchor,
                                      torch::Tensor positive,
                                      torch::Tensor negative,
                                      const float margin);
"""

# Compile the inline CUDA extension
triplet_ext = load_inline(
    name="triplet_margin_loss_ext",
    cpp_sources=cpp_source,
    cuda_sources=cuda_source,
    functions=["triplet_margin_loss_cuda"],
    verbose=False,
    extra_cflags=["-O3"],
    extra_cuda_cflags=["-O3", "--use_fast_math"],
)

# ------------------------------------------------------------------
# Optimized model - same public interface as the original Model
# ------------------------------------------------------------------
class ModelNew(nn.Module):
    def __init__(self, margin: float = 1.0):
        super().__init__()
        self.margin = float(margin)

    def forward(self, anchor: torch.Tensor, positive: torch.Tensor, negative: torch.Tensor) -> torch.Tensor:
        # Verify that all inputs are CUDA tensors on the same device
        if not (anchor.is_cuda and positive.is_cuda and negative.is_cuda):
            raise RuntimeError("ModelNew expects all inputs to be CUDA tensors")
        if not (anchor.device == positive.device == negative.device):
            raise RuntimeError("All inputs must be on the same CUDA device")

        # Ensure contiguous layout for raw pointer access in the kernel
        anchor   = anchor.contiguous()
        positive = positive.contiguous()
        negative = negative.contiguous()

        # Custom CUDA implementation returns a 0-dim tensor (scalar loss)
        return triplet_ext.triplet_margin_loss_cuda(
            anchor, positive, negative, self.margin
        )
\end{lstlisting}

\subsubsection{KL-Divergence Loss}
\noindent\textbf{KernelBench task:} 1-98\quad\textbf{Measured speedup:} $2.99\times$
\kzCodeLabel{Torch Module:}
\begin{lstlisting}[style=kzcode]
import torch
import torch.nn as nn

class Model(nn.Module):
    """
    A model that computes Kullback-Leibler Divergence for comparing two distributions.

    Parameters:
        None
    """
    def __init__(self):
        super(Model, self).__init__()

    def forward(self, predictions, targets):
        return torch.nn.functional.kl_div(torch.log(predictions), targets, reduction='batchmean')

batch_size = 8192 * 2
input_shape = (8192 * 2,)
dim = 1

def get_inputs():
    scale = torch.rand(())
    return [(torch.rand(batch_size, *input_shape)*scale).softmax(dim=-1), torch.rand(batch_size, *input_shape).softmax(dim=-1)]

def get_init_inputs():
    return []
\end{lstlisting}
\kzCodeLabel{CUDA generated by KernelZero:}
\begin{lstlisting}[style=kzcode]
import torch
import torch.nn as nn
from torch.utils.cpp_extension import load_inline

# -------------------- CUDA kernel --------------------
cuda_source = r'''
#include <torch/extension.h>
#include <cuda_runtime.h>
#include <cmath>

#ifndef BLOCK_SIZE
#define BLOCK_SIZE 256
#endif

// ------------------------------------------------------------------
// Kernel: accumulate sum of KL-divergence contributions.
// ------------------------------------------------------------------
__global__ void kl_div_sum_kernel(const float* __restrict__ pred,
                                  const float* __restrict__ target,
                                  float* __restrict__ out,
                                  const int total_elems) {
    // Each thread processes a strided chunk of the tensors.
    float thread_sum = 0.0f;
    for (int idx = blockIdx.x * blockDim.x + threadIdx.x;
         idx < total_elems;
         idx += gridDim.x * blockDim.x) {
        float p = pred[idx];
        float t = target[idx];
        // Safe log for positive probabilities (inputs are softmaxed > 0).
        float logp = logf(p);
        float logt = logf(t);
        thread_sum += t * (logt - logp);
    }

    // ----- block-level reduction (shared memory) -----
    __shared__ float shm[BLOCK_SIZE];
    int lane = threadIdx.x;
    shm[lane] = thread_sum;
    __syncthreads();

    // Reduce within the block.
    for (int offset = blockDim.x >> 1; offset > 0; offset >>= 1) {
        if (lane < offset) {
            shm[lane] += shm[lane + offset];
        }
        __syncthreads();
    }

    // Thread 0 adds the block's partial sum to the global accumulator.
    if (lane == 0) {
        atomicAdd(out, shm[0]);
    }
}

// ------------------------------------------------------------------
// Kernel: apply the 'batchmean' divisor.
// ------------------------------------------------------------------
__global__ void kl_div_finalize_kernel(float* out, const int batch_size) {
    if (threadIdx.x == 0 && blockIdx.x == 0) {
        out[0] = out[0] / static_cast<float>(batch_size);
    }
}

// ------------------------------------------------------------------
// C++ wrapper exposed to Python.
// ------------------------------------------------------------------
torch::Tensor kl_div_cuda(torch::Tensor pred, torch::Tensor target) {
    TORCH_CHECK(pred.is_cuda() && target.is_cuda(),
                "predictions and targets must be CUDA tensors");
    TORCH_CHECK(pred.dtype() == torch::kFloat && target.dtype() == torch::kFloat,
                "only float32 tensors are supported");

    const int total = pred.numel();          // batch * dim
    const int batch = pred.size(0);          // batch dimension

    // Allocate a single-element tensor for the result and zero-initialize it.
    auto out = torch::zeros({1}, pred.options());

    const int block = BLOCK_SIZE;
    const int grid = (total + block - 1) / block;   // may be > 65535 - supported on modern GPUs

    // 1) Sum contributions.
    kl_div_sum_kernel<<<grid, block>>>(
        pred.data_ptr<float>(),
        target.data_ptr<float>(),
        out.data_ptr<float>(),
        total
    );

    // 2) Divide by batch size (reduction='batchmean').
    kl_div_finalize_kernel<<<1, 1>>>(out.data_ptr<float>(), batch);

    // Return a scalar tensor (no extra dimensions).
    return out.squeeze();
}
'''

# -------------------- C++ interface --------------------
cpp_source = r'''
torch::Tensor kl_div_cuda(torch::Tensor pred, torch::Tensor target);
'''

# Compile the inline extension.
kl_div_mod = load_inline(
    name="kl_div_extension",
    cpp_sources=cpp_source,
    cuda_sources=cuda_source,
    functions=["kl_div_cuda"],
    verbose=False,
    extra_cflags=["-O3"],
    extra_cuda_cflags=["-O3", "--use_fast_math"],
)

# -------------------- Optimized model --------------------
class ModelNew(nn.Module):
    def __init__(self):
        super().__init__()

    def forward(self, predictions, targets):
        """
        predictions : tensor of shape (batch, D) - already softmaxed (probabilities)
        targets     : tensor of shape (batch, D) - already softmaxed (probabilities)
        Returns scalar KL-divergence with reduction='batchmean'.
        """
        return kl_div_mod.kl_div_cuda(predictions, targets)
\end{lstlisting}

\subsubsection{Fused GEMM, Divide, Sum, and Scaling}
\noindent\textbf{KernelBench task:} 2-14\quad\textbf{Measured speedup:} $20.11\times$
\kzCodeLabel{Torch Module:}
\begin{lstlisting}[style=kzcode]
import torch
import torch.nn as nn

class Model(nn.Module):
    """
    Model that performs a matrix multiplication, division, summation, and scaling.
    """
    def __init__(self, input_size, hidden_size, scaling_factor):
        super(Model, self).__init__()
        self.weight = nn.Parameter(torch.randn(hidden_size, input_size))
        self.scaling_factor = scaling_factor

    def forward(self, x):
        """
        Args:
            x (torch.Tensor): Input tensor of shape (batch_size, input_size).
        Returns:
            torch.Tensor: Output tensor of shape (batch_size, hidden_size).
        """
        x = torch.matmul(x, self.weight.T)  # Gemm
        x = x / 2  # Divide
        x = torch.sum(x, dim=1, keepdim=True) # Sum
        x = x * self.scaling_factor  # Scaling
        return x

batch_size   = 1024  
input_size   = 8192  
hidden_size  = 8192 
scaling_factor = 1.5

def get_inputs():
    return [torch.rand(batch_size, input_size)]

def get_init_inputs():
    return [input_size, hidden_size, scaling_factor]
\end{lstlisting}
\kzCodeLabel{CUDA generated by KernelZero:}
\begin{lstlisting}[style=kzcode]
import torch
import torch.nn as nn
from torch.utils.cpp_extension import load_inline

# -------------------- CUDA source --------------------
cuda_source = r"""
#include <torch/extension.h>
#include <cuda_runtime.h>

#define BLOCK_SIZE 256

// Kernel 1: compute column-wise sum of weight matrix (hidden x input)
//          weight is stored row-major (hidden * input)
__global__ void col_sum_kernel(const float* __restrict__ weight,
                               float* __restrict__ col_sum,
                               int hidden, int input) {
    int col = blockIdx.x * BLOCK_SIZE + threadIdx.x;
    if (col >= input) return;

    float sum = 0.0f;
    // stride over rows (hidden dimension)
    for (int row = 0; row < hidden; ++row) {
        sum += weight[row * input + col];
    }
    col_sum[col] = sum;
}

// Kernel 2: for each batch row compute
//          out = scaling_factor * (0.5 * sum_k x_k * col_sum_k)
__global__ void fused_kernel(const float* __restrict__ x,
                             const float* __restrict__ col_sum,
                             float* __restrict__ out,
                             int batch, int input,
                             float scaling_factor) {
    int b = blockIdx.x;                     // one block per batch element
    if (b >= batch) return;

    const float* x_row = x + b * input;
    __shared__ float shmem[BLOCK_SIZE];

    float acc = 0.0f;
    // each thread processes a strided slice of the input vector
    for (int k = threadIdx.x; k < input; k += BLOCK_SIZE) {
        acc += x_row[k] * col_sum[k];
    }

    // reduction within the block
    shmem[threadIdx.x] = acc;
    __syncthreads();

    // tree reduction
    for (int stride = BLOCK_SIZE / 2; stride > 0; stride >>= 1) {
        if (threadIdx.x < stride) {
            shmem[threadIdx.x] += shmem[threadIdx.x + stride];
        }
        __syncthreads();
    }

    // thread 0 writes the final scalar (batch, 1)
    if (threadIdx.x == 0) {
        // divide by 2 as required by the original model
        float val = shmem[0] * 0.5f * scaling_factor;
        out[b] = val;          // out is (batch, 1) stored row-major
    }
}

// C++ wrapper
torch::Tensor fused_forward(torch::Tensor x,
                            torch::Tensor weight,
                            double scaling_factor) {
    const int batch = x.size(0);
    const int input = x.size(1);
    const int hidden = weight.size(0);   // (hidden, input)

    // allocate temporary column-sum tensor
    auto col_sum = torch::empty({input}, x.options());
    // allocate output (batch, 1)
    auto out = torch::empty({batch, 1}, x.options());

    // launch column-sum kernel
    const dim3 block1(BLOCK_SIZE);
    const dim3 grid1((input + BLOCK_SIZE - 1) / BLOCK_SIZE);
    col_sum_kernel<<<grid1, block1>>>(
        weight.const_data_ptr<float>(),
        col_sum.mutable_data_ptr<float>(),
        hidden,
        input);

    // launch fused kernel
    const dim3 block2(BLOCK_SIZE);
    const dim3 grid2(batch);
    fused_kernel<<<grid2, block2>>>(
        x.const_data_ptr<float>(),
        col_sum.const_data_ptr<float>(),
        out.mutable_data_ptr<float>(),
        batch,
        input,
        static_cast<float>(scaling_factor));

    return out;
}
"""

# -------------------- C++ source --------------------
cpp_source = r"""
torch::Tensor fused_forward(torch::Tensor x,
                            torch::Tensor weight,
                            double scaling_factor);
"""

# Compile the inline extension
_fused_mod = load_inline(
    name="fused_gemm_sum",
    cpp_sources=cpp_source,
    cuda_sources=cuda_source,
    functions=["fused_forward"],
    verbose=False,
    extra_cflags=["-O3"],
    extra_cuda_cflags=["-O3", "--use_fast_math"],
)

# -------------------- Optimized Model --------------------
class ModelNew(nn.Module):
    def __init__(self, input_size: int, hidden_size: int, scaling_factor: float) -> None:
        super().__init__()
        # keep exactly the same parameter as the original model
        self.weight = nn.Parameter(torch.randn(hidden_size, input_size))
        self.scaling_factor = scaling_factor

    def forward(self, x):
        """
        x : Tensor of shape (batch, input_size) on CUDA
        returns Tensor of shape (batch, 1) on CUDA
        """
        return _fused_mod.fused_forward(x, self.weight, self.scaling_factor)
\end{lstlisting}

\subsubsection{Fused Transposed Convolution and Global Pooling}
\noindent\textbf{KernelBench task:} 2-44\quad\textbf{Measured speedup:} $2.50\times$
\kzCodeLabel{Torch Module:}
\begin{lstlisting}[style=kzcode]
import torch
import torch.nn as nn

class Model(nn.Module):
    """
    Model that performs a transposed convolution, multiplies by a scalar, applies global average pooling, 
    another global average pooling
    """
    def __init__(self, in_channels, out_channels, kernel_size, stride, padding, output_padding, multiplier):
        super(Model, self).__init__()
        self.conv_transpose = nn.ConvTranspose2d(in_channels, out_channels, kernel_size, stride=stride, padding=padding, output_padding=output_padding)
        self.multiplier = multiplier

    def forward(self, x):
        x = self.conv_transpose(x)
        x = x * self.multiplier
        x = torch.mean(x, dim=[2, 3], keepdim=True)  # First global average pooling
        x = torch.mean(x, dim=[2, 3], keepdim=True)  # Second global average pooling
        return x

batch_size = 16
in_channels = 64
out_channels = 128
height, width = 128, 128
kernel_size = 3
stride = 2
padding = 1
output_padding = 1
multiplier = 0.5

def get_inputs():
    return [torch.rand(batch_size, in_channels, height, width)]

def get_init_inputs():
    return [in_channels, out_channels, kernel_size, stride, padding, output_padding, multiplier]
\end{lstlisting}
\kzCodeLabel{CUDA generated by KernelZero:}
\begin{lstlisting}[style=kzcode]
import torch
import torch.nn as nn
from torch.utils.cpp_extension import load_inline

# ------------------- CUDA source -------------------
cuda_source = r"""
#include <torch/extension.h>
#include <cuda_runtime.h>

#define BLOCK_SIZE 256

// Compute per (batch, out_channel) the sum of a ConvTranspose2d output,
// multiply by a scalar, and produce a (1,1) spatial average.
// The final output shape is (N, C_out, 1, 1).
__global__ void transposed_conv_avg_kernel(
    const float* __restrict__ input,      // (N, C_in, H_in, W_in)
    const float* __restrict__ weight,     // (C_in, C_out, kH, kW)
    const float* __restrict__ bias,       // (C_out)   - may be nullptr
    float* __restrict__ output,           // (N, C_out, 1, 1)
    const int N,
    const int C_in,
    const int C_out,
    const int H_in,
    const int W_in,
    const int kH,
    const int kW,
    const int stride,
    const int padding,
    const int output_padding,
    const float multiplier)
{
    // One block per (n, oc)
    const int block_id = blockIdx.x;
    const int n = block_id / C_out;
    const int oc = block_id % C_out;

    // Compute output spatial size (same as PyTorch ConvTranspose2d)
    const int H_out = (H_in - 1) * stride - 2 * padding + kH + output_padding;
    const int W_out = (W_in - 1) * stride - 2 * padding + kW + output_padding;

    // Factor that will be applied after the spatial average.
    const float factor = multiplier / (float)(H_out * W_out);

    // Each thread processes a slice of the input channels.
    const int tid = threadIdx.x;
    const int stride_threads = BLOCK_SIZE;

    float thread_sum = 0.0f;

    // Loop over input channels.
    for (int c_in = 0; c_in < C_in; ++c_in) {
        // Compute sum of all H*W elements for this (n, c_in)
        float in_sum = 0.0f;
        const int base_in = ((n * C_in + c_in) * H_in + 0) * W_in;
        const int num_elems = H_in * W_in;
        for (int idx = tid; idx < num_elems; idx += stride_threads) {
            in_sum += input[base_in + idx];
        }

        // Compute sum of the kernel slice for (c_in, oc)
        float w_sum = 0.0f;
        // weight layout: (C_in, C_out, kH, kW) - contiguous in last dims
        const int w_base = ((c_in * C_out + oc) * kH) * kW;
        for (int k = 0; k < kH * kW; ++k) {
            w_sum += weight[w_base + k];
        }

        // Accumulate contribution.
        thread_sum += in_sum * w_sum;
    }

    // Reduce thread_sum across the block.
    extern __shared__ float shmem[];
    shmem[tid] = thread_sum;
    __syncthreads();

    // Parallel reduction (power-of-two block size assumed)
    for (unsigned int s = BLOCK_SIZE / 2; s > 0; s >>= 1) {
        if (tid < s) {
            shmem[tid] += shmem[tid + s];
        }
        __syncthreads();
    }

    // Thread 0 writes the final result for this (n, oc)
    if (tid == 0) {
        // Bias contribution: each output pixel receives bias[oc]
        float bias_term = 0.0f;
        if (bias != nullptr) {
            bias_term = bias[oc] * (float)(H_out * W_out);
        }
        float total = (shmem[0] + bias_term) * factor;   // (N, C_out, 1, 1) value
        output[n * C_out + oc] = total;
    }
}

// C++ wrapper
torch::Tensor transposed_conv_avg_cuda(
    torch::Tensor input,
    torch::Tensor weight,
    torch::Tensor bias,
    const double multiplier,
    const int stride,
    const int padding,
    const int output_padding,
    const int kernel_size)
{
    const int N   = input.size(0);
    const int C_in = input.size(1);
    const int H_in = input.size(2);
    const int W_in = input.size(3);
    const int C_out = weight.size(1);
    const int kH = kernel_size;
    const int kW = kernel_size;

    // Allocate output (N, C_out, 1, 1)
    auto output = torch::empty({N, C_out, 1, 1}, input.options());

    const int blocks = N * C_out;
    const dim3 grid(blocks);
    const dim3 block(BLOCK_SIZE);
    const size_t shared_mem_bytes = BLOCK_SIZE * sizeof(float);

    // Launch kernel
    transposed_conv_avg_kernel<<<grid, block, shared_mem_bytes>>>(
        input.const_data_ptr<float>(),
        weight.const_data_ptr<float>(),
        bias.defined() ? bias.const_data_ptr<float>() : nullptr,
        output.mutable_data_ptr<float>(),
        N, C_in, C_out,
        H_in, W_in,
        kH, kW,
        stride, padding, output_padding,
        static_cast<float>(multiplier));

    return output;
}
"""

cpp_source = r"""
torch::Tensor transposed_conv_avg_cuda(
    torch::Tensor input,
    torch::Tensor weight,
    torch::Tensor bias,
    const double multiplier,
    const int stride,
    const int padding,
    const int output_padding,
    const int kernel_size);
"""

# Compile the inline extension
transposed_conv_module = load_inline(
    name="transposed_conv_avg",
    cpp_sources=cpp_source,
    cuda_sources=cuda_source,
    functions=["transposed_conv_avg_cuda"],
    verbose=False,
    extra_cflags=["-O3"],
    extra_cuda_cflags=["-O3", "--use_fast_math"],
)

# ------------------- Optimized Model -------------------
class ModelNew(nn.Module):
    def __init__(self, in_channels, out_channels, kernel_size,
                 stride, padding, output_padding, multiplier):
        super(ModelNew, self).__init__()
        # Keep the original submodule so that state_dict keys stay identical.
        self.conv_transpose = nn.ConvTranspose2d(
            in_channels, out_channels, kernel_size,
            stride=stride, padding=padding,
            output_padding=output_padding
        )
        self.multiplier = multiplier

    def forward(self, x):
        # Call the handcrafted CUDA kernel.
        return transposed_conv_module.transposed_conv_avg_cuda(
            x,
            self.conv_transpose.weight,
            self.conv_transpose.bias,
            self.multiplier,
            self.conv_transpose.stride[0],
            self.conv_transpose.padding[0],
            self.conv_transpose.output_padding[0],
            self.conv_transpose.kernel_size[0]
        )
\end{lstlisting}